\pdfoutput=1
\documentclass[10pt,twocolumn,letterpaper]{article}

\usepackage{cvpr}
\usepackage{times}
\usepackage{epsfig}
\usepackage{epstopdf}
\usepackage{graphicx}
\usepackage{amsmath}
\usepackage{amssymb}
\usepackage{subfigure}
\usepackage{psfrag}

\newcommand{\bs}{\boldsymbol}
\newcommand{\tf}{{\rm f}}
\newcommand{\R}{\mathbb{R}}

\usepackage[pagebackref=true,breaklinks=true,letterpaper=true,colorlinks,bookmarks=false]{hyperref}
\def\equationautorefname~#1\null{Eq.~(#1)\null}

\makeatletter
\renewcommand{\maketag@@@}[1]{\hbox{\m@th\normalsize\normalfont#1}}%
\makeatother

\cvprfinalcopy 

\def\cvprPaperID{536} 

\ifcvprfinal\fi
\begin{document}

\title{An Efficiently Coupled Shape and Appearance Prior for Active Contour Segmentation}

\author{Martin Mueller\\
Rivian Automotive LLC.\\
607 Hansen Way, Palo Alto, CA 94304, USA\\
{\tt\small martin.mueller@gatech.edu}
\and
Navdeep Dahiya and Anthony Yezzi \\
Georgia Institute of Technology, \\
85 Fifth Street NW, Atlanta, GA 30332, USA\\
{\tt\small \{ndahiya3@, ayezzi@ece.\}gatech.edu}
}

\maketitle

\begin{abstract}
This paper proposes a novel training model based on shape and appearance features for object segmentation in images and videos. Whereas most such models rely on two-dimensional appearance templates or a finite set of descriptors, our appearance-based feature is a one-dimensional function, which is efficiently coupled with the object's shape by integrating intensities along the object's iso-contours. Joint PCA training on these shape and appearance features further exploits shape-appearance correlations and the resulting training model is incorporated in an active-contour-type energy functional for recognition-segmentation tasks. Experiments on synthetic and infrared images demonstrate how this shape and appearance training model improves accuracy compared to methods based on the Chan-Vese energy.
\end{abstract}

\section{Introduction}
Identifying the precise location and extent of an object in an image is an important pre-processing step for many computer vision applications, such as recognition or tracking. Whereas generic segmentation methods partition an image into regions based on low-level image statistics, prior knowledge about the objects of interest may improve segmentation performance in the presence of noise, occlusion, and model errors. A review of these \textit{recognition}-segmentation approaches with different priors, such as color, motion, shape, and texture is given in \cite{cremers2007review}. Shape is a powerful feature for many applications, particularly for medical imaging where shapes of body parts do not vary much among patients. Shape models are plentiful and range from statistical models~\cite{heimann2009statistical} to landmark points, whose local modes of variation are computed from a training set~\cite{cootes1995active}, to Fourier-domain representations of boundaries~\cite{staib1992boundary}. Most relevant to our work are level set representations of shape, which haven been incorporated as shape priors for active contour techniques, \eg in~\cite{bresson2006variational, chen2001incorporation, leventon2000statistical, rousson2002shape,  tsai2003shape}.   

Additional robustness can be achieved when combining shape with appearance. For example,~\cite{cootes2001active} extends the landmark approach in \cite{cootes1995active} by learned and deformable appearance templates. Examples of level set approaches are \cite{fritscher20073d, huang2008metamorphs, yang20043d} where principal component analysis (PCA) is performed on the level sets and on the pixel-intensity image of the training set to obtain coupled shape and appearance features. Our approach is similar to these methods in that level set shape and intensity are jointly trained on using PCA. However, we propose a smaller, transformed set of intensity features (a one-dimensional function) by numerically integrating image intensities along iso-contours of the object's shape. In this way, our appearance feature is---by definition and even before training---coupled to the shape of an object. Moreover, this one-dimensional appearance feature does not require computationally expensive two-dimensional warps of intensity templates to migrate between different shape configurations. Our method is a compromise between two-dimensional appearance templates with their high computational complexity and finite number of appearance statistics (such as intensity mean or variance) with their limited discrimination power.


The remainder of the paper is organized as follows. \autoref{sec:models} introduces the shape and appearance training models. In particular, \autoref{sec:modelsS} borrows a shape training model from a previous publication, \autoref{sec:modelsA} proposes our new and efficient appearance descriptor, and \autoref{sec:modelsSA} couples the two models through joint training. These priors are incorporated into a recognition-segmentation energy in \autoref{sec:energy}, which is minimized by gradient descent. \autoref{sec:experiments} demonstrates the algorithm's performance in comparison to the method in~\cite{tsai2003shape} on synthetic and infrared images.

\section{Shape and Appearance Training Models \label{sec:models}}
Shape is an intrinsic feature for structured objects and is not affected by illumination, which makes it a robust feature. However, shape information can be corrupted, for example in the presence of occlusions, and extracting the correct shape is not an easy task. In such cases, it becomes essential to consider photometric features as well. The fact that geometric and photometric properties are intimately coupled through the physical object that creates the image motivates their joint exploitation. In the following, two discriminators, a geometric (shape) and a photometric (appearance) one, will be presented, which are then coupled through joint training to exploit correlations between them.

\subsection{Shape-Based Training Model \label{sec:modelsS}}
Training models for shape are obtained as explained in more detail in~\cite{tsai2003shape} by PCA on aligned level set representations of the training shapes. By ``shape'' we mean the boundary curve $C$ of an object in the image, which is represented as the zero level set of the corresponding (unique) $L^2$-signed-distance level set function $\psi$, i.e., $\psi(x)=\pm \min_{y\in C} \| x - y \|_2$, where $\psi(x)$ is positive if $x$ is outside of $C$ (background), and negative inside (object). Given $N$ training shapes, they are first aligned to maximize the mutual total overlap. Then, after PCA on the aligned shapes, the level set shape model $\Phi(x;\bs w)$ is written as
\begin{equation}
\Phi(x;\bs w) = \bar{\Phi}(x) + \sum_{i=1}^K w_i \Phi_i(x),
\label{eq:shapePCAmodel}
\end{equation}
where $\bar{\Phi}(x) $ is the mean shape, $\Phi_i(x),\,i=1,...,K, \,K<N$ are the $K$ eigenshapes with the largest singular values, and $\bs w=(w_1,...,w_K)$ are free shape parameters. Note that none of $\Phi(x;\bs w)$, $\bar{\Phi}(x)$ or $\Phi_i(x)$ are signed-distance functions in general.

\subsection{Appearance-Based Training Model \label{sec:modelsA}}
\begin{figure}
\centering
\subfigure[level set curves $C_T$]{\epsfig{figure=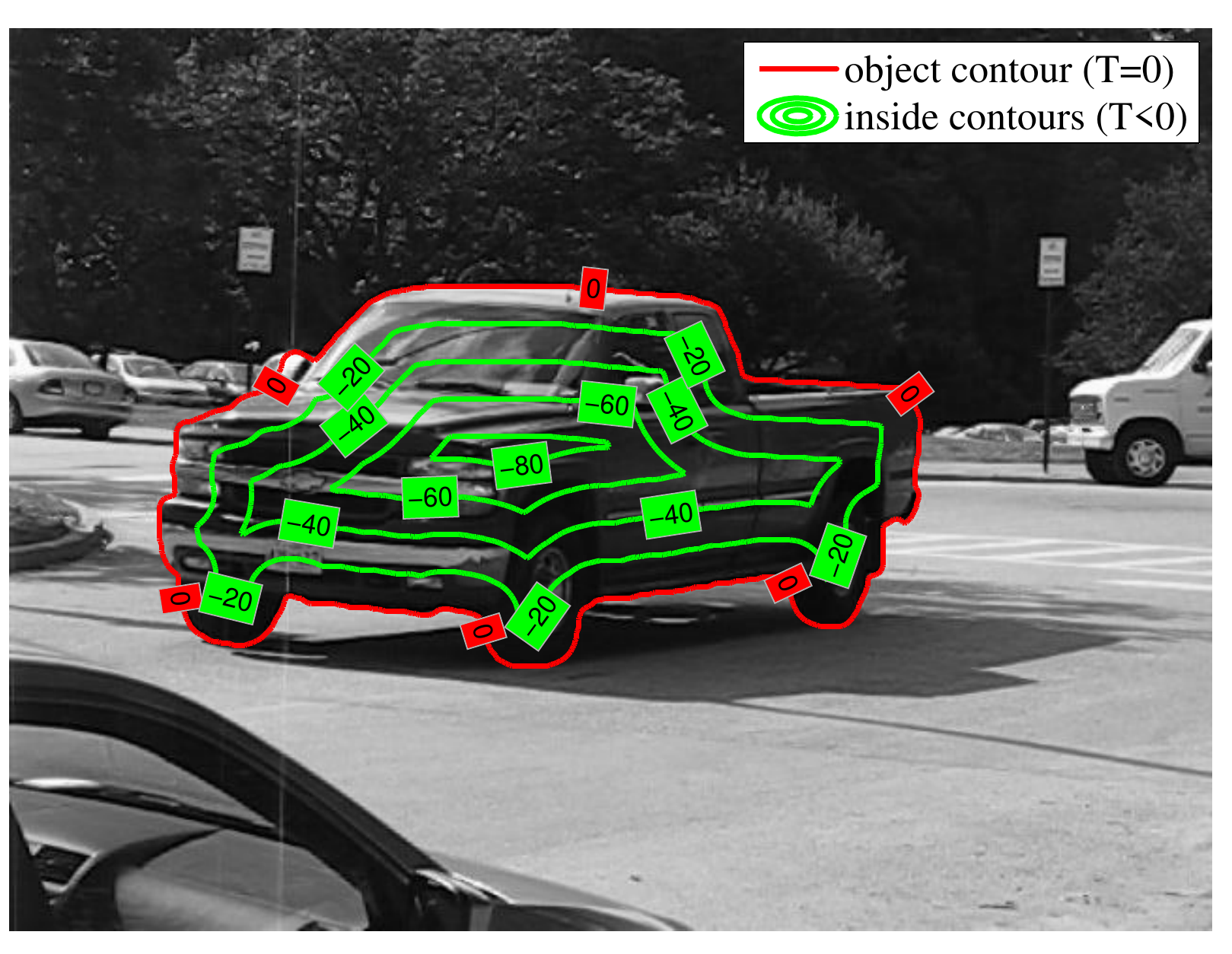, width=0.40\textwidth} \label{fig:f_T_levelsets}} 
\subfigure[mean intensities along $T$-level-sets]{
\psfrag{barfT}{\footnotesize $\text{f}(T)$}
\psfrag{T}{\raisebox{-0.5em}{\footnotesize $T$}}
\psfrag{100}{\hspace{-1pt}\scriptsize 100}
\psfrag{90}{}\psfrag{80}{}\psfrag{70}{}\psfrag{60}{}
\psfrag{50}{\hspace{-1pt}\scriptsize 50}
\psfrag{40}{}\psfrag{30}{}\psfrag{20}{}
\psfrag{-80}{\raisebox{-0.5em}{\footnotesize -80}}
\psfrag{-60}{\raisebox{-0.5em}{\footnotesize -60}}
\psfrag{-20}{\raisebox{-0.5em}{\footnotesize -20}}
\psfrag{0}{\raisebox{-0.5em}{\footnotesize 0}}
\psfrag{-70}{}\psfrag{-50}{}\psfrag{-40}{}\psfrag{-30}{}\psfrag{-10}{}
\epsfig{figure=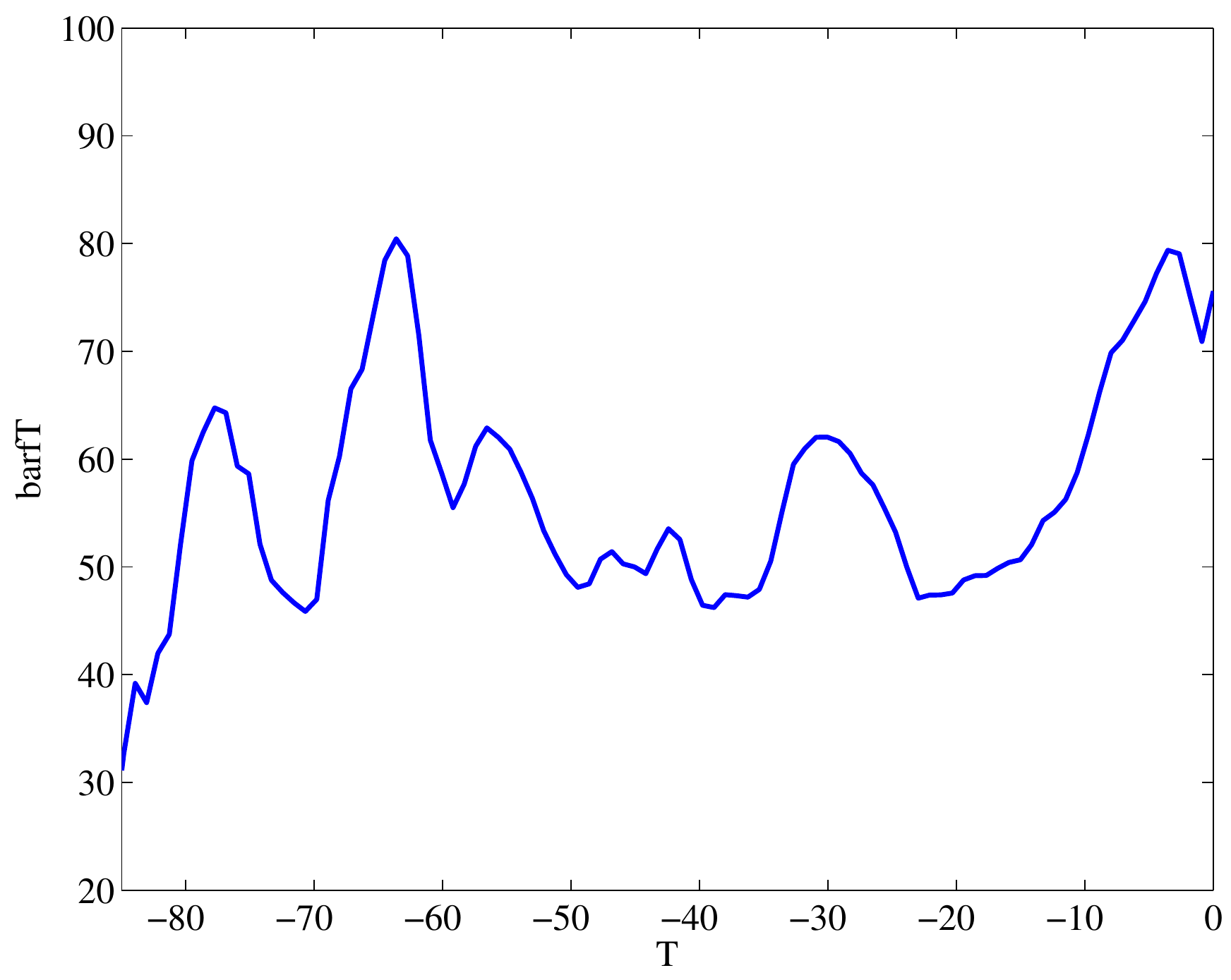, width=0.45\textwidth, height=0.15\textwidth} \label{fig:fT_fT}}
\caption{Illustration of the photo-geometric representation.}
\label{fig:f_T}
\end{figure}

Two-dimensional templates are a popular choice as appearance features, but computationally complex. Instead, given an image $I$ and the object boundary shape $C$ (and its equivalent signed-distance function $\psi$), we propose as an appearance-based discriminator the mean image intensities along iso-contours of the object shape, i.e., our one-dimensional template is defined as 
\begin{equation}
\text{f}(T) = \frac{\int_{C_T}I\, ds} {\text{length}(C_T)},
\label{eq:fT2}
\end{equation}
where $\tf: [\psi_\text{min},\, 0] \mapsto \R$ are the mean intensities and $C_T = \{x: \psi(x)=T \}$ are the iso-contours. In \autoref{fig:f_T}, these quantities are illustrated for the image of a truck from the Berkeley Motion Segmentation dataset~\cite{brox2010object}. The level sets $C_T$ for some values of $T$ between $\psi_\text{min}$ and $0$ are shown in \autoref{fig:f_T_levelsets} and the mean intensity along these curves is plotted in \autoref{fig:fT_fT}. For numerical integration on level sets, see, for example, \cite{min2007geometric}. For a more efficient technique for integration along several level sets in one shot, see \cite{dahiya2021intrinsic}.

We call $\text{f}(T)$ the \textit{photo-geometric} representation of an object because it couples the object's geometric information $C_T$ with its photometric information $I$. As a feature, it is well-suited for training purposes because it is invariant to translation and rotation. It is also invariant to scale if the domain of $\tf(T)$ is normalized to a constant interval, \eg, $[-1,\, 0]$. As an appearance feature, the photo-geometric representation can be viewed as a compromise between the competing goals of robustness, efficiency, and discrimination power. Active appearance models~\cite{cootes2001active} are a powerful, yet cumbersome method for appearance modeling. On the other end, there are simple methods, in which object intensity is modeled simply as a constant, \eg, Chan-Vese~\cite{chan2001active}, or more generally through a finite set of statistics. While more robust and tractable, these methods may not be accurate enough to describe more complicated object appearance. Our proposed algorithm using the one-dimensional photo-geometric descriptor is more general than finite numbers of statistics, and at the same time not as cumbersome as two-dimensional template models.

Analogous to the shape-based training model, PCA is performed on the photo-geometric representations $\text{f}_i(T),\,i=1,...,N$ of $N$ training samples to obtain an appearance-based training model from this training set. Owing to the photo-geometric representation being invariant to translation and rotation, alignment is not needed. The resulting appearance-based training model $\text{F}(\tau;\bs v)$ is 
\begin{equation}
\text{F}(\tau;\bs v) = \bar{\text{F}}(\tau) + \sum_{i=1}^L v_i \text{F}_i(\tau),
\label{eq:appPCAmodel}
\end{equation}
where $\bar{\text{F}}$ is the mean photo-geometric representation, $\text{F}_i$ are the eigencomponents, and $\bs v=(v_1,...,v_L)$ are free appearance parameters. Switching from $T$ to $\tau$ is to indicate that the domains of all training samples need to be scaled to the same interval, e.g., to $\tau \in [-1,\, 0]$.

\subsection{Coupled Training Model \label{sec:modelsSA}}
To exploit both shape and appearance, shape and appearance PCA can be performed independently leading to a shape and appearance training model with $K+L$ free parameters. Alternatively, by combining shape and appearance in one PCA, an equally good or better training model with less parameters is expected since correlations between shape and appearance can be exploited. The coupling is achieved by performing PCA on the Cartesian product of shape $\psi$ and photo-geometric representation $\text{f}(T)$ of objects, that is by stacking the shape vector and the radiance vector into one column vector $\Lambda_i$ and performing PCA on these stacked vectors $\Lambda_1,...,\Lambda_N$ . The resulting coupled training model is
\begin{equation}
\Lambda(x,\tau;\bs w) = \bar{\Lambda}(x,\tau) + \sum_{i=1}^M w_i \Lambda_i(x,\tau),
\end{equation}
which yields after de-stacking $\Lambda(x,\tau;\bs w)$
\begin{subequations}
\begin{align}
\Phi^\text{cpl}(x;\bs w) &= \bar{\Phi}^\text{cpl}(x) + \sum_{i=1}^M w_i \Phi_i^\text{cpl}(x), \\
\text{F}^\text{cpl}(\tau;\bs w) &= \bar{\text{F}}^\text{cpl}(\tau) + \sum_{i=1}^M w_i \text{F}_i^\text{cpl}(\tau),
\end{align}
\label{eq:TMcoupled}%
\end{subequations}
where $\Lambda(x,\tau;\bs w) = (\Phi^\text{cpl}(x;\bs w), \text{F}^\text{cpl}(\tau;\bs w))$. Note that the eigenvectors and means in this coupled case (indicated by superscript ``cpl'') are different from the decoupled case.

\section{Active Contour Energy with Priors \label{sec:energy}}
In the following, a recognition-segmentation active contour energy is defined and minimized, which incorporates the shape and appearance training models from \autoref{sec:models}. Analogous to \cite{tsai2003shape}, object shape is restricted to the linear training model and the energy is minimized with respect to the shape model parameters $\bs w$. Moreover, the energy compares the input image with the learned appearance-based training model in the least squares sense, attempting to find a good match between the data and the model by finding optimal model weights $\bs v$. In the case of a coupled shape and appearance training model, the additional condition $\bs w=\bs v$ applies. 

In addition, the energy depends on pose parameters $\bs p$, which determine the transformation that maps coordinates from the training domain (denoted by $x$) to the recognition domain of the input image (denoted by $\hat{x}$, i.e., $I = I(\hat{x})$). In general, this transformation $g$ acts as $\hat{x}  = g(x; \bs p)$. For this application, we choose $g$ to be a similarity transformation with its parameters $\bs p$ representing translation, rotation, and scale. The transformation from the training domain to the input image domain is then carried out as follows
\begin{equation}
\hat{\Phi}(\hat{x};\bs w, \bs p) = \Phi(g^{-1}(\hat{x};\bs p);\bs w)
\label{eq:hatphi}
\end{equation}
for the shape-based model. In the following, all quantities transformed in this fashion are indicated by the $\hat{(\cdot)}$ notation.

Finally, the recognition energy is defined as 
\begin{equation}
E(\bs w, \bs v, \bs p)=E^\text{in}(\bs w, \bs v, \bs p)+E^\text{out}(\bs w, \bs p)
\label{eq:E}
\end{equation}
 where
\begin{align}
&E^\text{in}(\bs w, \bs v, \bs p) = \int_{\hat{R}(\bs w, \bs p)}  \alpha \left (I-\text{F}(\hat{\Phi};\bs v) \right)^2 \nonumber \\ 
&\quad \quad \quad \quad \quad \quad \quad+ \beta  \left ( \frac{\partial\text{F} }{\partial\tau}(\hat{\Phi};\bs v) \| \nabla_{\hat{x}} \hat{\Phi} \|   \right )^2 d\hat{x}, \label{eq:Ein} \\
&E^\text{out}(\bs w, \bs p) = \int_{\hat{R}^c(\bs w, \bs p)}  \alpha  \left (I-u_\text{out}\right)^2 \, d\hat{x}
\end{align}
are region integrals modeling the object domain ($E^\text{in}$) and the background domain ($E^\text{out}$) of the image. Whereas the background term is simply a term from the Chan-Vese energy with shape prior and intensity mean $u_\text{out}$ as found in \cite{tsai2003shape} (because we assume that an appearance-based model for the background is not available in most applications), the object term $E^\text{in}$ incorporates the novel appearance-based training model \autoref{eq:appPCAmodel} proposed in this paper. The first term in \autoref{eq:Ein} (the fidelity term) measures the discrepancy between the image intensity $I$ and the appearance-based model $\text{F}(\hat{\Phi};\bs v)$ on the estimated object region $\hat{R}(\bs w, \bs p)=\{ x :\hat{\Phi}(\hat{x};\bs w, \bs p) < 0 \}$ weighted by a positive parameter $\alpha$. The second term in \autoref{eq:Ein} (a regularization term) penalizes the magnitude of the gradient of the appearance-based model in the two-dimensional image domain (as opposed to the derivative in the one-dimensional domain of the photo-geometric representation) weighted by another parameter $\beta$. Penalizing the two-dimensional gradient is critical to prevent the level set from arbitrarily shrinking to lower the cost. 

The recognition algorithm searches for the shape, appearance, and pose parameters minimizing the energy by gradient descent. The derivatives of $E^\text{out}$ with respect to the parameters have been shown in~\cite{tsai2003shape}. The computation of the derivatives of $E^\text{in}$ is provided in the supplementary material of this paper. Here, the resulting derivatives are simply stated as
\scriptsize
\begin{align}
&\frac{\partial E^\text{in}}{\partial p_i} = \int_{\hat{C}}  \left ( \alpha \left (I-\text{F}(0) \right)^2  + \beta  \left ( \frac{\partial\text{F} }{\partial\tau}(0) \| \nabla_{\hat{x}} \hat{\Phi} \|   \right )^2 \right) \left( \hat{N} \cdot \frac{\partial g}{\partial p_i} \right)  d\hat{s} \nonumber  \\
&+ 2 \int_{\hat{R}} \Biggl [  \frac{\partial \text{F}}{\partial \tau} (\hat{\Phi})  \left( \nabla_{\hat{x}} \hat{\Phi} \cdot \frac{\partial g}{\partial p_i} \right) \left ( \alpha \left (I-\text{F}(\hat{\Phi}) \right) - \beta   \frac{\partial^2\text{F} }{\partial\tau^2}(\hat{\Phi}) \| \nabla_{\hat{x}} \hat{\Phi} \| ^2  \right ) \nonumber \\
&- \beta   \left(\frac{\partial \text{F}}{\partial \tau}(\hat{\Phi})\right)^2 \nabla_{\hat{x}} \hat{\Phi} \cdot  \left( H_{\hat{x}}\hat{\Phi} \frac{\partial g}{\partial p_i} + \frac{\partial^2 g}{\partial x \partial p_i} \left [   \frac{\partial g}{\partial x} \right ]^{-1} \nabla_{\hat{x}} \hat{\Phi}  \right )  \Biggr ] d\hat{x}
\end{align}\normalsize
for the pose parameters, where $H_{\hat{x}}\hat{\Phi}$ denotes the Hessian of $\hat{\Phi}$,
\scriptsize
\begin{align}
&\frac{\partial E^\text{in}}{\partial w_i} =  -\int_{\hat{C}}  \left ( \alpha \left (I-\text{F}(0) \right)^2  + \beta  \left ( \frac{\partial\text{F} }{\partial\tau}(0) \| \nabla_{\hat{x}} \hat{\Phi} \|   \right )^2 \right) \frac{\hat{\Phi}_i}{\| \nabla_{\hat{x}} \hat{\Phi}  \|}  d\hat{s} \nonumber  \\
&+ 2 \int_{\hat{R}} \Biggl [ \beta  \left(\frac{\partial \text{F}}{\partial \tau}(\hat{\Phi})\right)^2  \hat{h}_i \cdot  \left( \left [   \frac{\partial g}{\partial x} \right ]^{-1} \nabla_{\hat{x}} \hat{\Phi}  \right )     \nonumber \\
&- \frac{\partial \text{F}}{\partial \tau} (\hat{\Phi}) \hat{\Phi}_i  \left ( \alpha \left (I-\text{F}(\hat{\Phi}) \right) - \beta   \frac{\partial^2\text{F} }{\partial\tau^2}(\hat{\Phi}) \| \nabla_{\hat{x}} \hat{\Phi} \| ^2  \right ) \Biggr ] d\hat{x}, \label{eq:Eams_w}
\end{align}\normalsize
for the shape parameters, where $h_i(x) = \nabla_x \Phi_i(x)$, and so  $\hat{h}_i$ is  $h_i$ transformed to the image domain analogous to \autoref{eq:hatphi}. Finally, 
\scriptsize
\begin{align}
\frac{\partial E^\text{in}}{\partial v_i} &=  2 \int_{\hat{R}} \alpha \left (\text{F}(\hat{\Phi})-I \right) \text{F}_i(\hat{\Phi}) + \beta \frac{\partial \text{F}}{\partial \tau} (\hat{\Phi}) \| \nabla_{\hat{x}} \hat{\Phi} \|^2 \frac{d \text{F}_i}{d \tau} (\hat{\Phi}) \, d\hat{x},
\label{eq:Eams_v}
\end{align}\normalsize
for the appearance parameters. The update equation for a certain parameter $a$ (where $a$ can be any of the shape, appearance or pose parameters) from step $t$ to step $t+1$ is then given by
\begin{align}
a^{t+1} = a^t - \delta t \, \left ( \frac{\partial E^\text{in}}{\partial a}+\frac{\partial E^\text{out}}{\partial a} \right )
\end{align}
where $\delta t$ is a sufficiently small positive step size. 

The above equations assume that shape and appearance PCAs were performed independently. If shape and appearance are coupled according to \autoref{eq:TMcoupled}, then, due to the fact that shape and appearance are controlled by the same parameter, \autoref{eq:Eams_w} and \autoref{eq:Eams_v} are added to form the derivative with respect to the coupled shape and appearance parameters.

\section{Experiments \label{sec:experiments}}

Several experiments are presented in this section to demonstrate the algorithm's mechanics as well as its advantages over the Chan-Vese energy with shape priors \cite{tsai2003shape} and without priors \cite{chan2001active}. First, two synthetic experiments are presented to analyze the algorithm. Then three tracking results in infrared videos are presented. Infrared imagery presents a promising field of application for the region-based active contour methods compared in this paper, since infrared images are smoother and less rich in local feature detail compared to images in the visible spectrum, so that standard feature point methods are less readily applicable. The following abbreviations will be used throughout this section:
\begin{itemize}
\itemsep0em
\item CV: Chan-Vese without shape prior \cite{chan2001active},
\item CV-S: Chan-Vese with shape prior \cite{tsai2003shape},
\item E-SAd: \autoref{eq:E} with decoupled priors \autoref{eq:shapePCAmodel} and \textcolor{red}{(\ref{eq:appPCAmodel})},
\item E-SAc: \autoref{eq:E} with coupled priors \autoref{eq:TMcoupled},
\item E-SA:  \autoref{eq:E} with either coupled or decoupled priors.
\end{itemize}

\begin{figure*}
\centering
\subfigure[training objects]{
\epsfig{figure=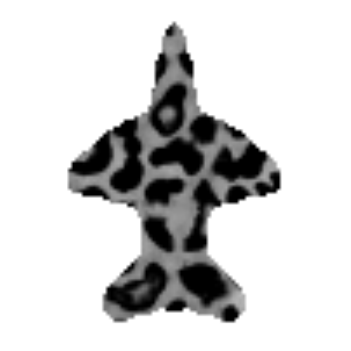,width=0.076\textwidth}
\epsfig{figure=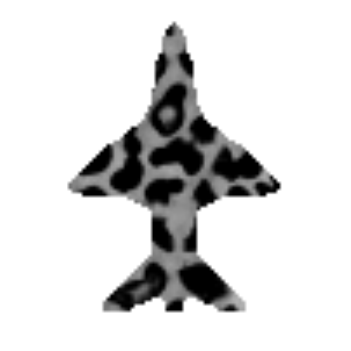,width=0.076\textwidth}
\epsfig{figure=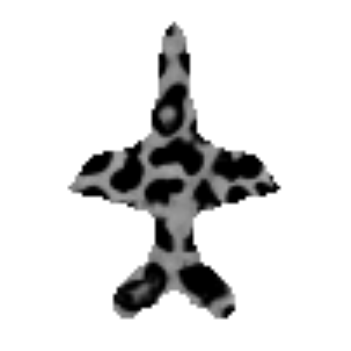,width=0.076\textwidth}
\epsfig{figure=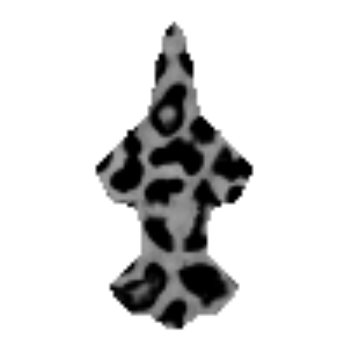,width=0.076\textwidth}
\epsfig{figure=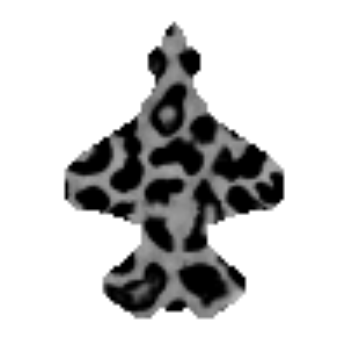,width=0.076\textwidth}
\epsfig{figure=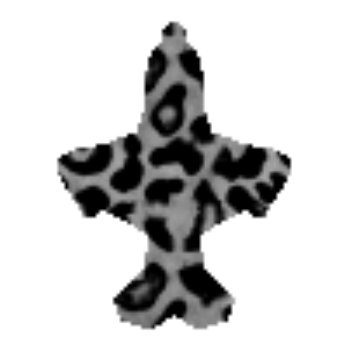,width=0.076\textwidth}
\epsfig{figure=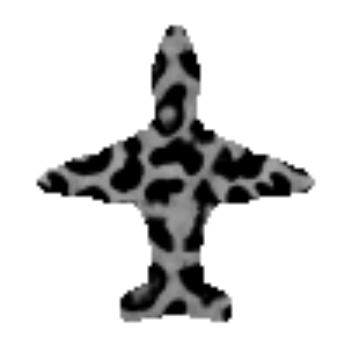,width=0.076\textwidth}
\epsfig{figure=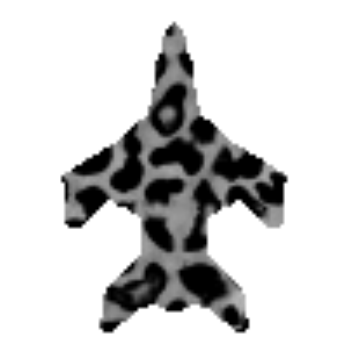,width=0.076\textwidth}
\epsfig{figure=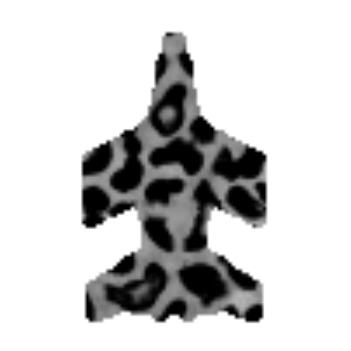,width=0.076\textwidth}
\epsfig{figure=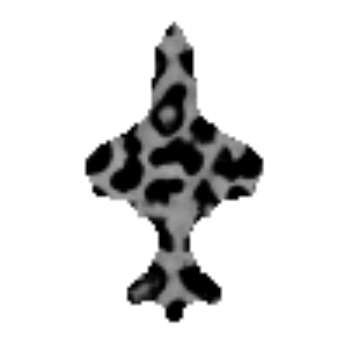,width=0.076\textwidth}
\epsfig{figure=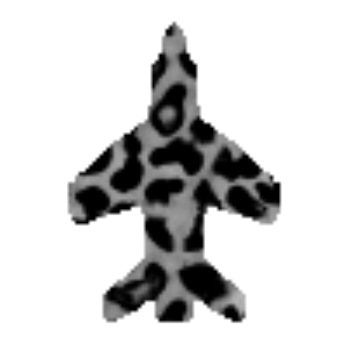,width=0.076\textwidth}
\epsfig{figure=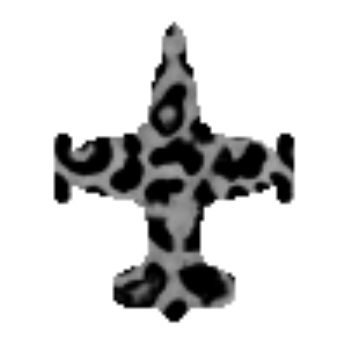,width=0.076\textwidth}\label{fig:exp1_trainingdata}} \\
\centering
\subfigure[initialization]{
\epsfig{figure=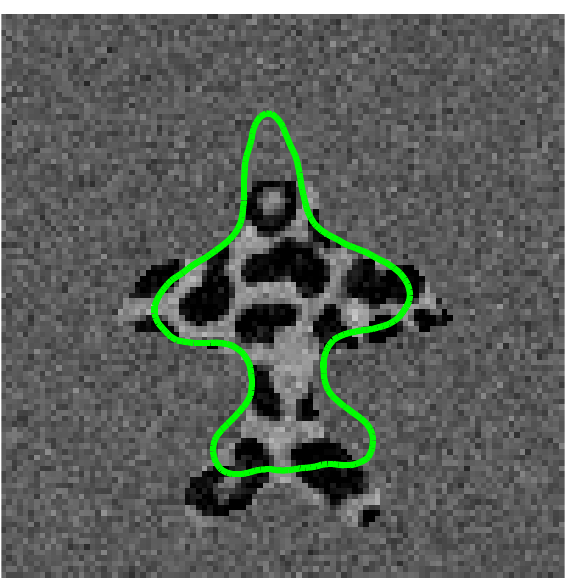,width=0.16\textwidth}\label{fig:fighters_initial}\label{fig:exp1_testimage}
}
\subfigure[CV]{
\epsfig{figure=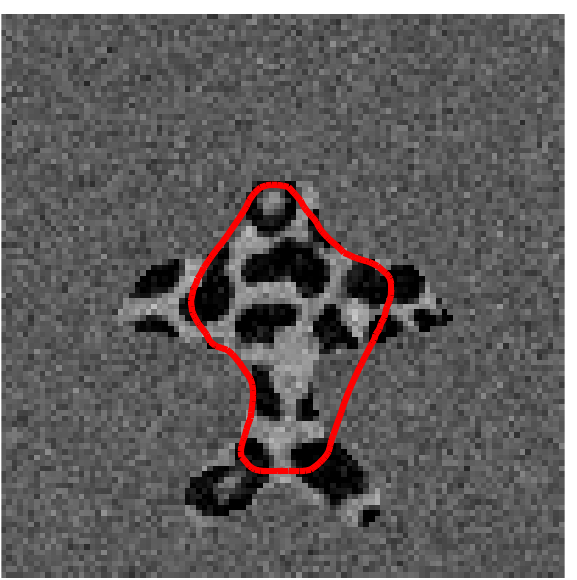,width=0.16\textwidth}\label{fig:exp1_CV}
}
\subfigure[CV-S]{
\epsfig{figure=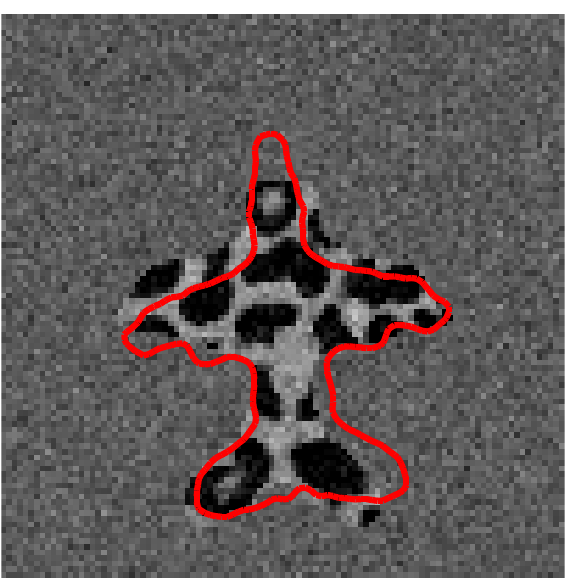,width=0.16\textwidth}\label{fig:exp1_CVS}
}
\subfigure[E-SAd]{
\epsfig{figure=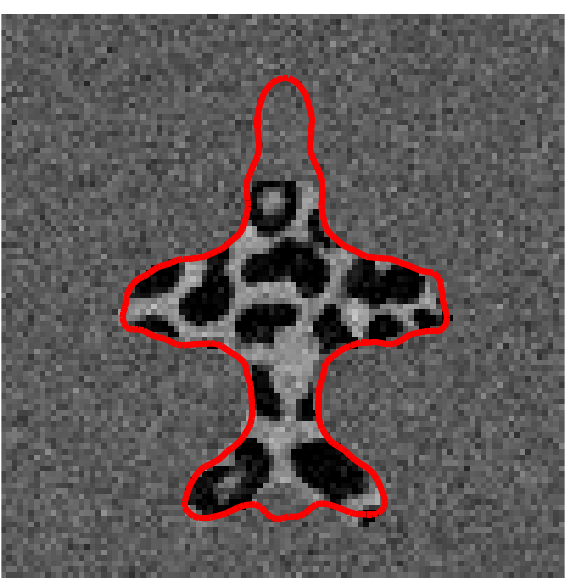,width=0.16\textwidth}\label{fig:exp1_EASd}
}
\subfigure[E-SAc]{
\epsfig{figure=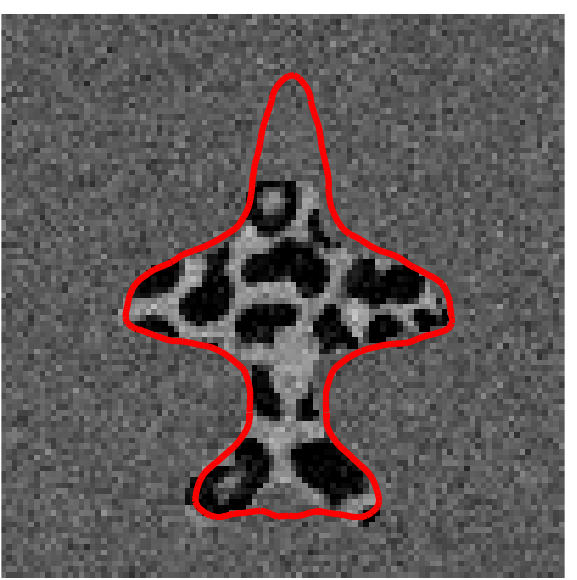,width=0.16\textwidth}\label{fig:exp1_EASc}
}
\caption{Experiment 1 (Fighters): (a) training samples to create shape and apperance models, (b) test image and initial segmentation (mean shape with offset pose parameters), (c)-(f) final segmentation results using different algorithms. The proposed algorithms E-SAd and E-SAc have a clear advantage over the CV methods.}
\label{fig:fighters}
\end{figure*}

\medskip
\noindent \textbf{Experiment 1 (Fighters).} In the first experiment, we compute training models for the fighter shapes with a leopard texture shown in \autoref{fig:exp1_trainingdata} and keep 4 eigencomponents from PCA. Note that, even though the leopard pattern is the same for all fighter shapes, the photo-geometric representation (by its definition) is not the same because shape varies among the samples. The synthetic test image in \autoref{fig:exp1_testimage} is created by superposing one training sample on a gray background image, then over-painting the top 30\% of the image with background gray-level, and finally adding zero-mean Gaussian noise with variance 15. These last two operations simulate occlusion and noise. The initial segmentation using the mean training shape and appearance as well as offset pose parameters is also shown in \autoref{fig:exp1_testimage}.

The localization results using CV, CV-S, E-SAd, and E-SAc are shown in \autoref{fig:exp1_CV} to \ref{fig:exp1_EASc}. Clearly, CV with no shape prior is an over-simplified model for this scenario. CV-S performs much better, indicating that shape priors are an important addition, but CV-S gets attracted to a local minimum. Both E-SAd and E-SAc are fairly accurate in recovering the global shape and most of its details. It cannot be expected to recover all details of shape since the PCA models use only a certain number of principal components so that---even though the test image is part of the training data---not all of the test image's information is present in the training model. Computation of the energy values for each gradient descent iteration reveals that E-SAd converges to a smaller minimum energy value than E-SAc, which is expected   because E-SAd is the more general model. On the other hand, the visual result is not necessarily better, which speaks in favor of E-SAc using only half as many shape and appearance parameters as E-SAd.

\begin{figure*}
\centering
 \subfigure[original image]{\epsfig{figure=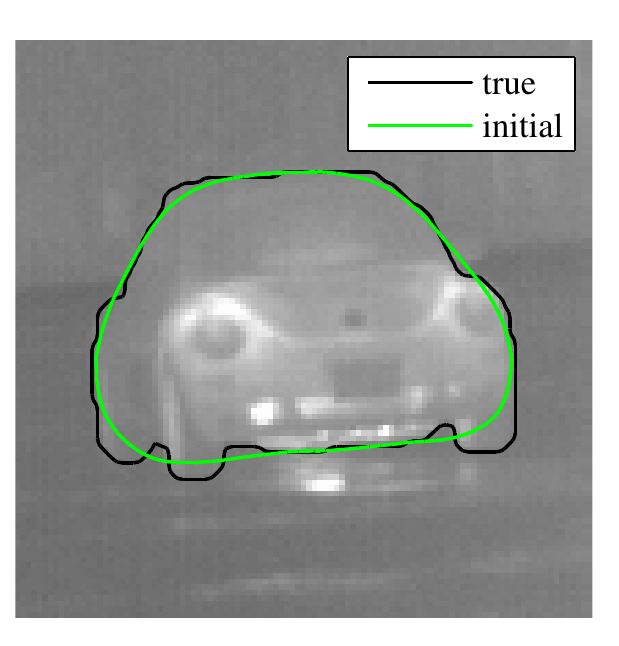,width=0.17\textwidth}\label{fig:exp2_original}} 
 \subfigure[$K=3$]{\epsfig{figure=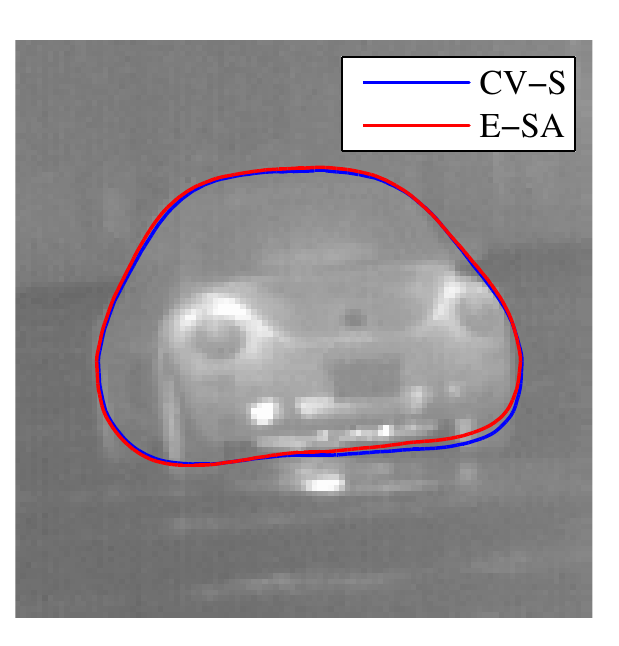,width=0.17\textwidth}\label{fig:exp2_segm_K3}}
 \subfigure[$K=10$]{\epsfig{figure=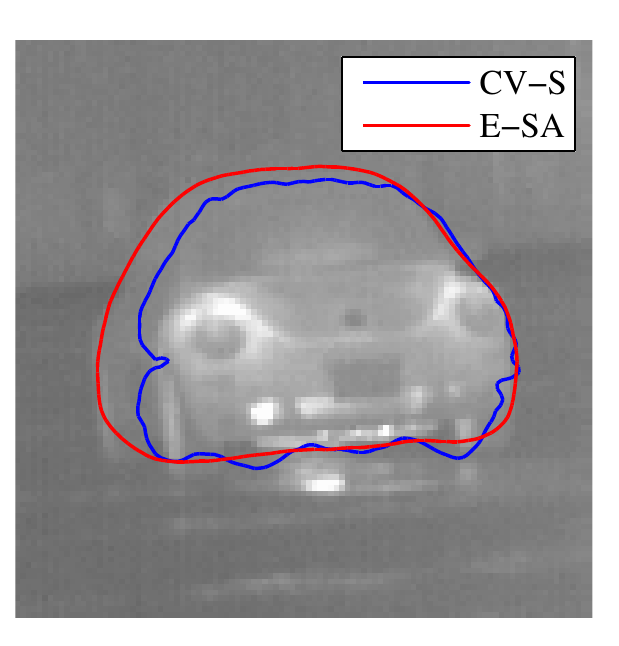,width=0.17\textwidth}\label{fig:exp2_segm_K10}}
 \subfigure[$K=30$]{\epsfig{figure=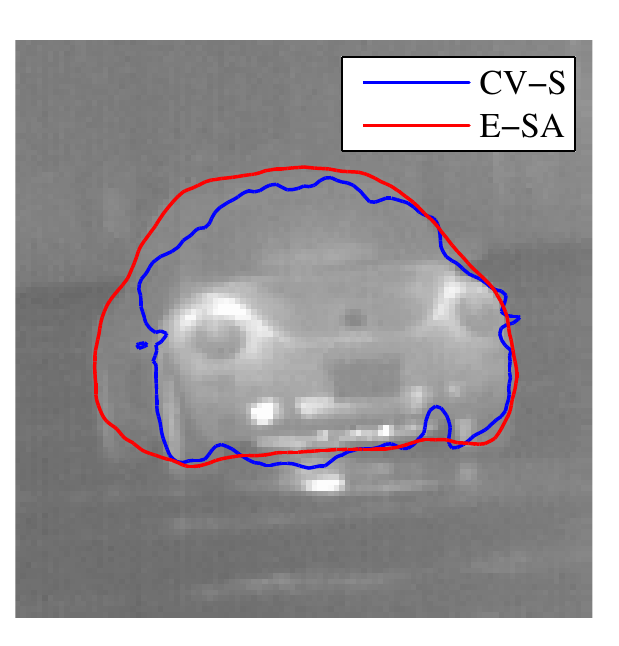,width=0.17\textwidth}\label{fig:exp2_segm_K30}} 
 \subfigure[$K=60$]{\epsfig{figure=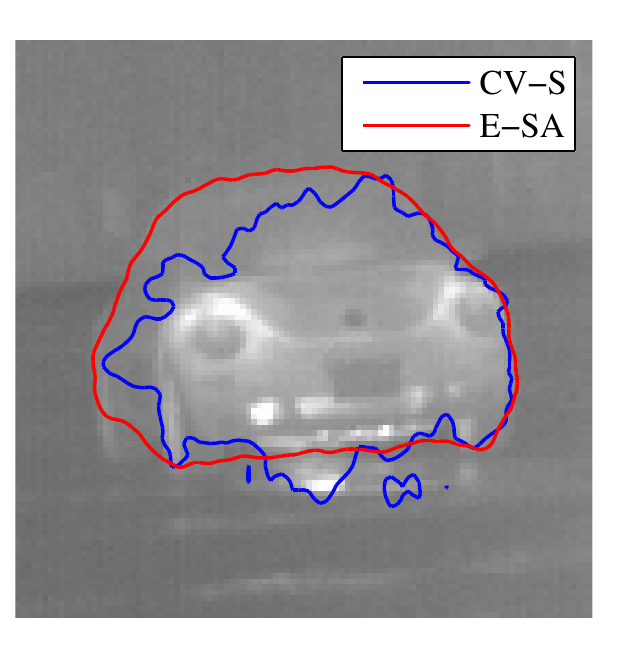,width=0.17\textwidth}\label{fig:exp2_segm_K60}}\\
 \subfigure[segmentation error]{\epsfig{figure=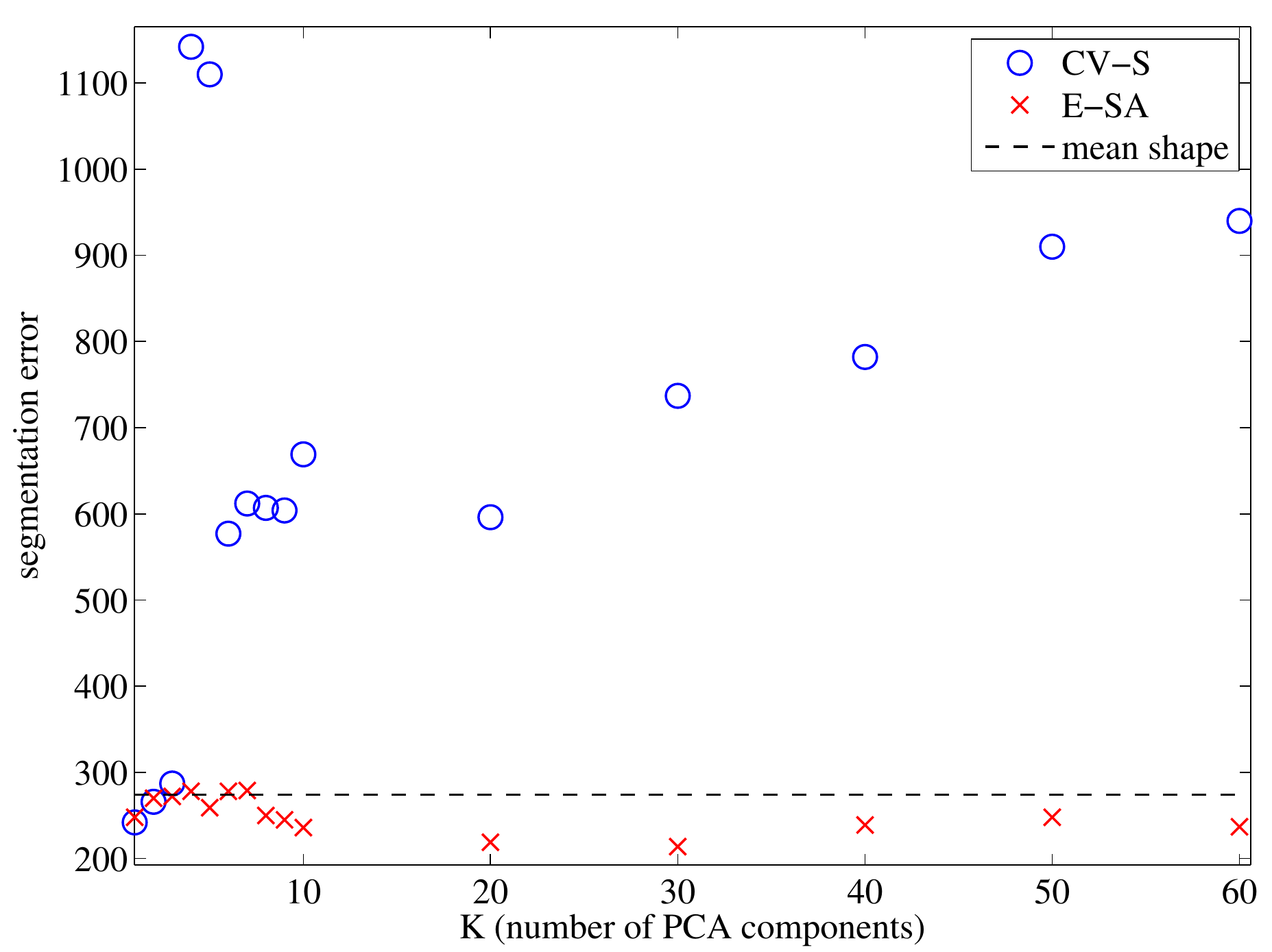,width=0.45\textwidth}\label{fig:exp2_ErroverK}} 
 \subfigure[minimum energy value]{\epsfig{figure=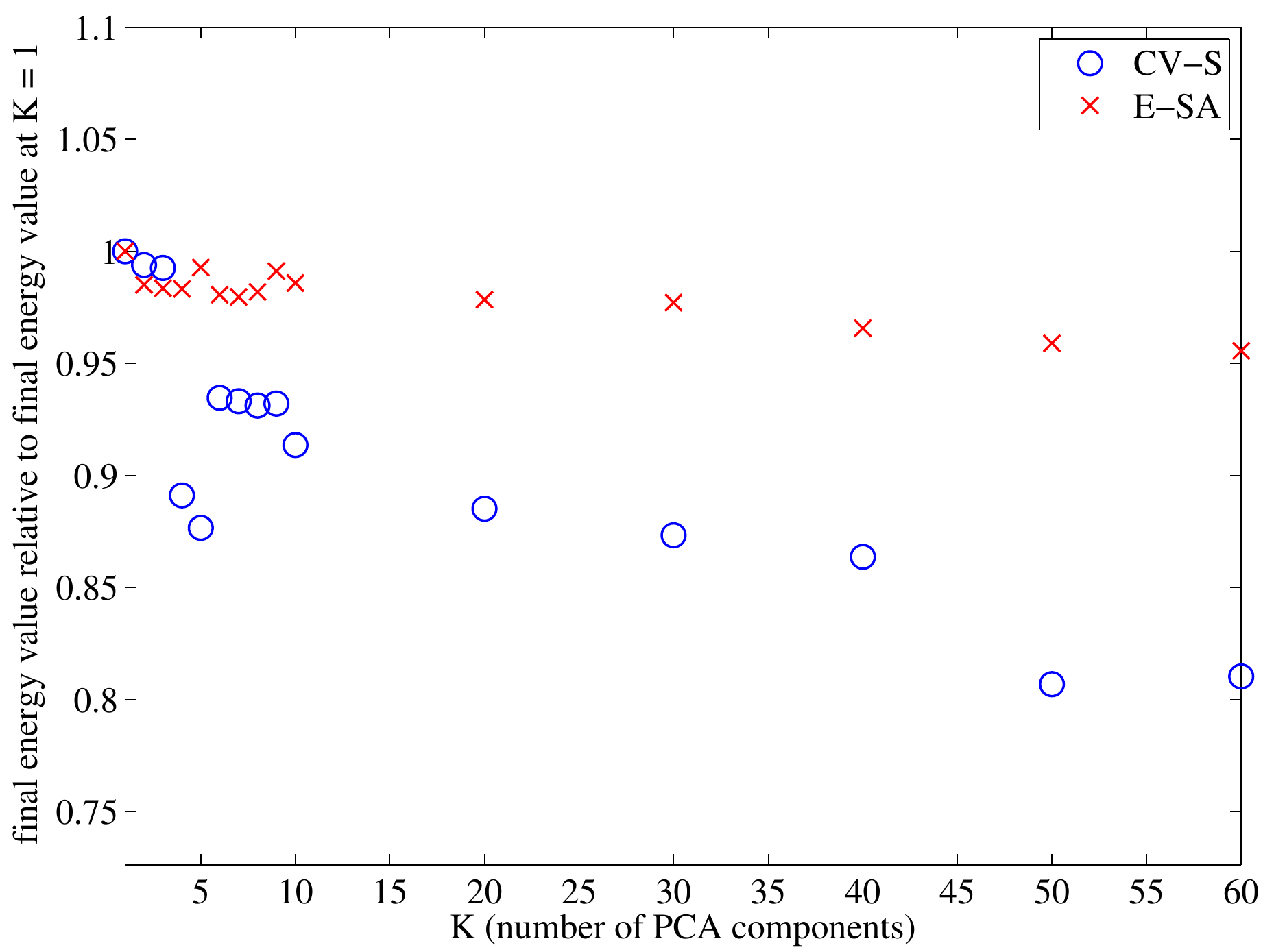,width=0.45\textwidth}\label{fig:exp2_EoverK}}
\caption{Experiment 2 (Beetle): (a) true shape (black) and mean shape from PCA (green), (b)-(e) final segmentation results for CV-S and E-SA with increasing number of eigencomponents $K$. Too many degrees of freedom make CV-S unstable, whereas for E-SA, accuracy increases as shown in (f). (g) illustrates that CV-S minimizes its energy, but that the scene is too complex for the CV energy since segmentation error actually increases.}
\label{fig:beetle_train}
\end{figure*}

\medskip
\noindent \textbf{Experiment 2 (Beetle).} The following experiment further demonstrates superior robustness of E-SA over CV-S. Consider the infrared image of a beetle from~\cite{morris2007statistics} in \autoref{fig:exp2_original} with true segmentation given by the black curve. A noisy shape-based model is obtained by shifting and rotating the true shape in different combinations and then performing shape PCA on the dislocated and rotated shapes. The resulting mean shape is used as initialization in \autoref{fig:exp2_original}.

In this experiment, we take the appearance-based model for E-SA  to be the photo-geometric representation of the original object (this would be equivalent to performing PCA on identical appearance vectors), and for CV-S, the true intensity mean of the object is computed and fixed in the CV-S algorithm. Then, $K$, the number of eigenshapes used in the shape-based model, is varied and the segmentation results are compared. Note that a larger $K$ is expected to produce more accurate results, since more information about the true object shape will be contained in the training model for larger $K$.

Visual results are shown in \autoref{fig:exp2_segm_K3} to \ref{fig:exp2_segm_K60} for CV-S in blue and E-SA in red. Whereas for $K=3$, the results are very similar to the mean shape, the larger $K$ the more CV-S deteriorates and the more accurate E-SA becomes. This observation is confirmed in \autoref{fig:exp2_ErroverK} where the segmentation error (integral of the square of the difference between the true and estimated object region) is plotted for various values of $K$. E-SA tends to slightly decrease segmentation error for larger $K$, whereas CV-S clearly increases the error and is not robust for $K> 3$. In \autoref{fig:exp2_EoverK}, the final minimum energy value for a certain $K$ is shown in relation to the final minimum energy value when $K=1$. Interestingly, it is observed that the final energy value for CV-S decreases faster for larger $K$ than it does for E-SA. This observation indicates that CV-S achieves its goal of minimizing the Chan-Vese energy correctly, but that the Chan-Vese energy is not a good model for this scenario since the actual segmentation error increases, whereas E-SA approaches the global minimum.


\begin{figure*}
\centering
\subfigure[E-SAc for Terravic video irw07]{
  \epsfig{figure=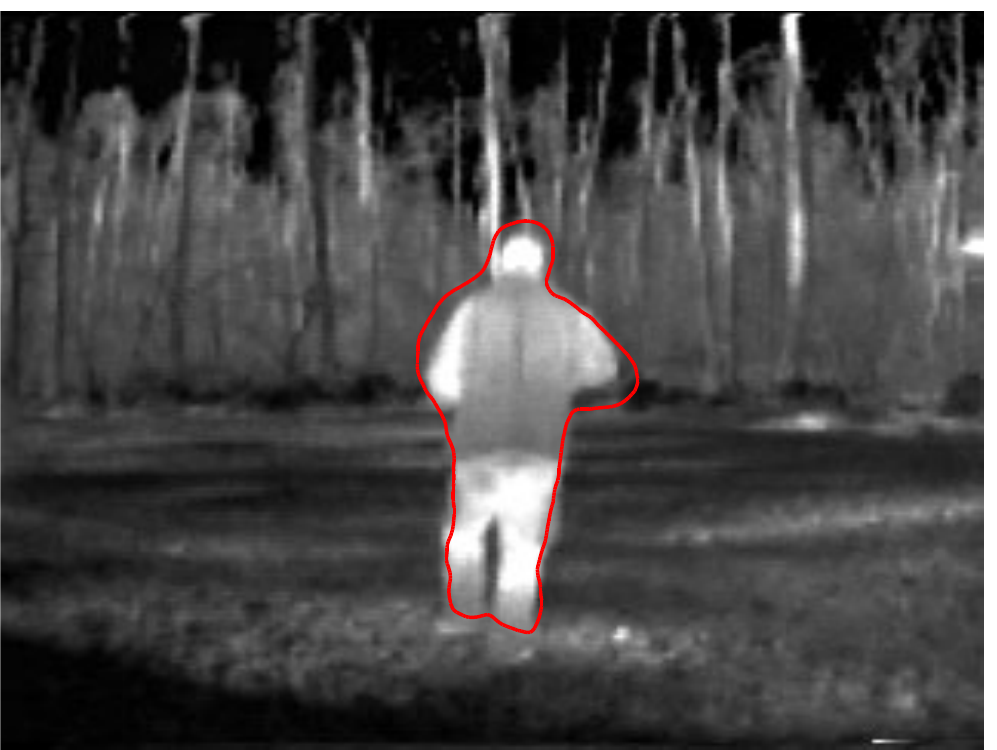,width=0.175\textwidth}
  \epsfig{figure=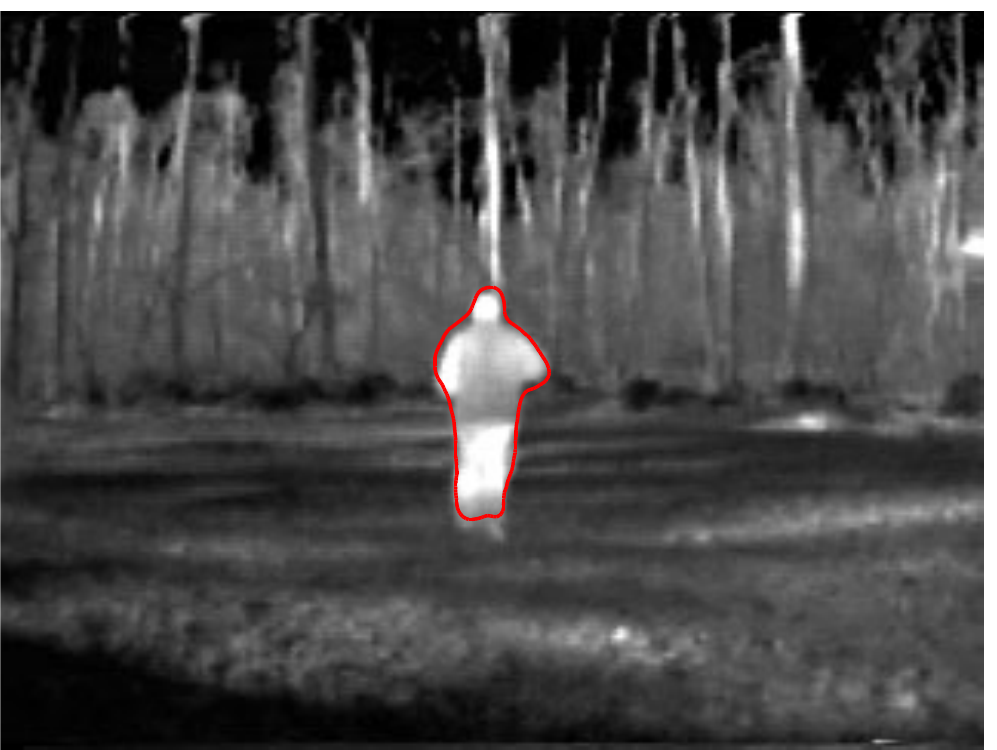,width=0.175\textwidth}
  \epsfig{figure=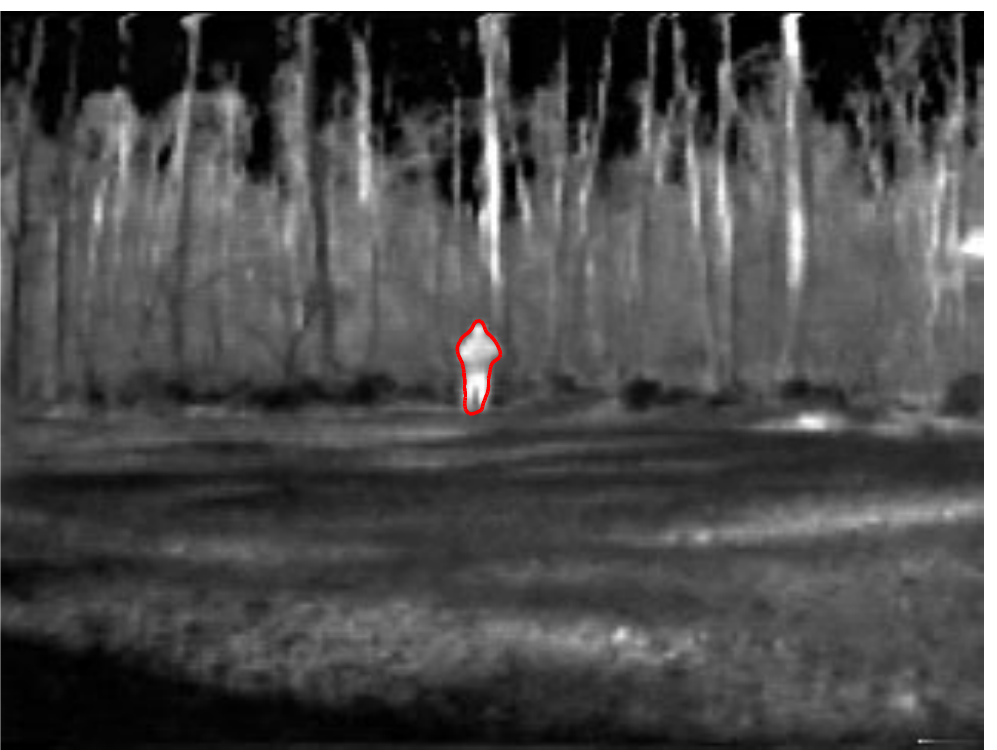,width=0.175\textwidth}
  \epsfig{figure=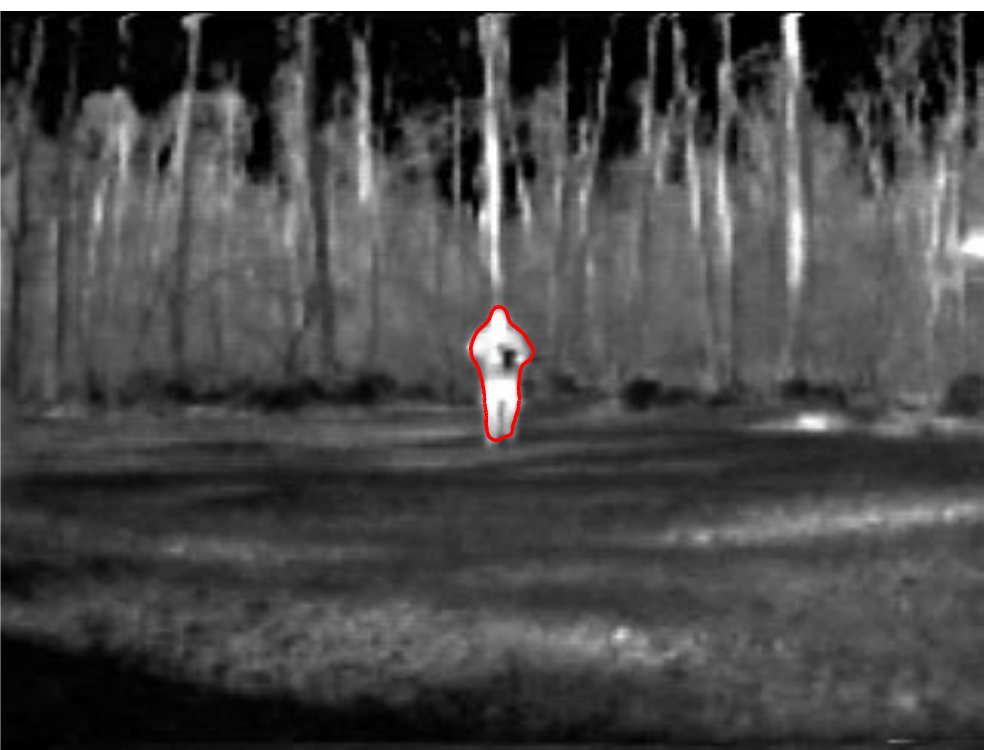,width=0.175\textwidth}
  \epsfig{figure=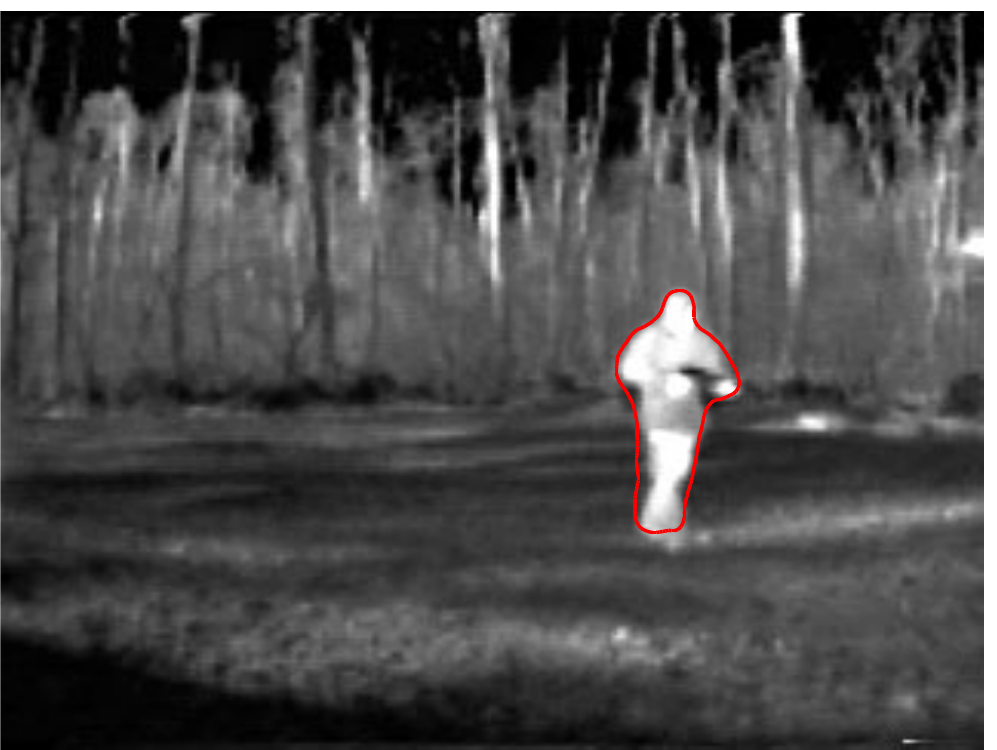,width=0.175\textwidth}
  \label{fig:exp3_irw07ESA}
}
\subfigure[CV-S for Terravic video irw07]{
  \epsfig{figure=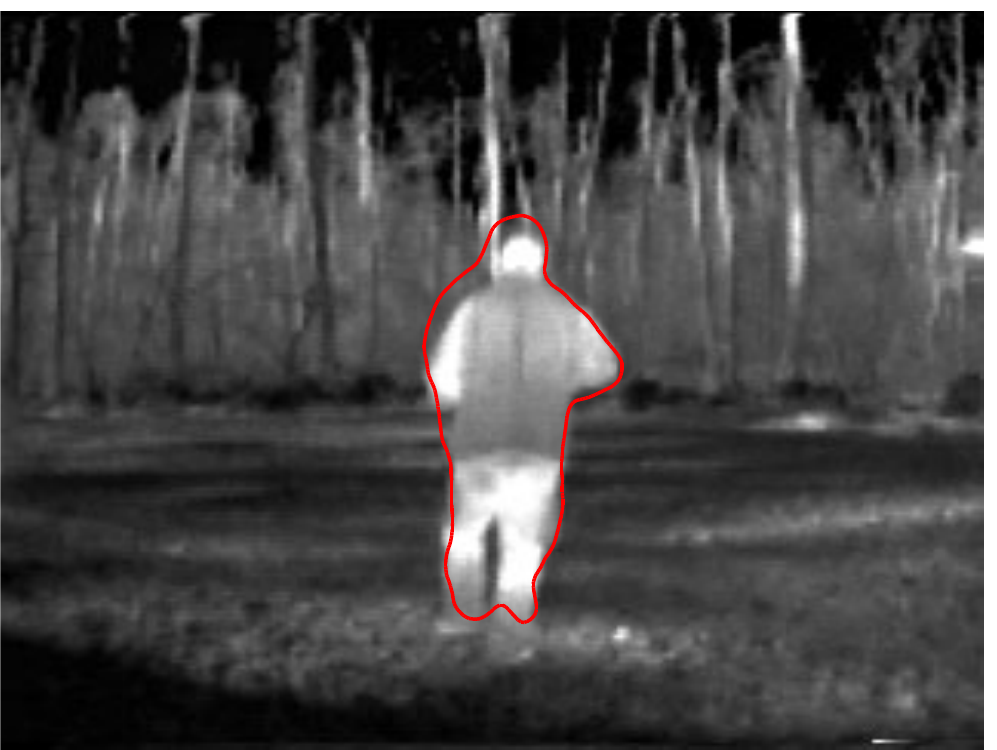,width=0.175\textwidth}
  \epsfig{figure=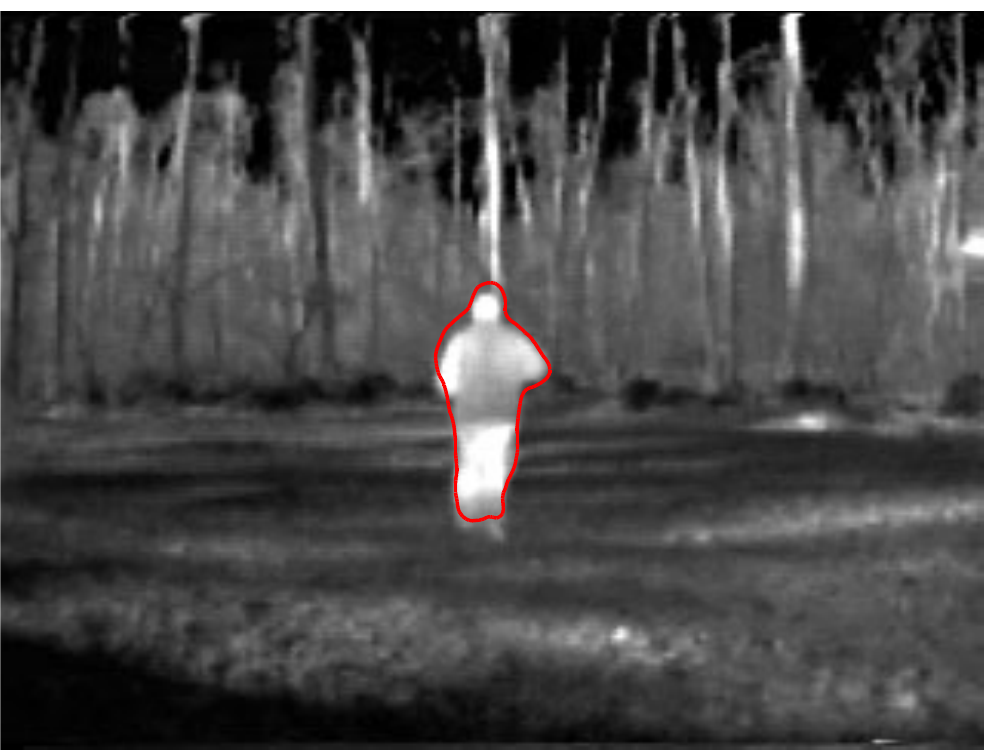,width=0.175\textwidth}
  \epsfig{figure=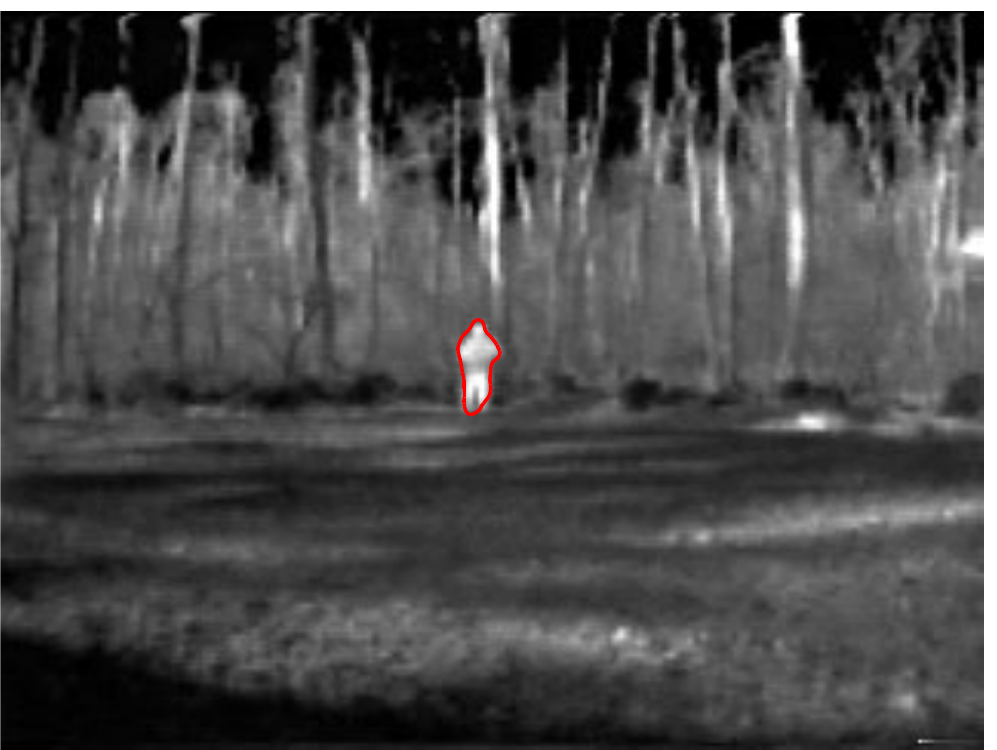,width=0.175\textwidth}
  \epsfig{figure=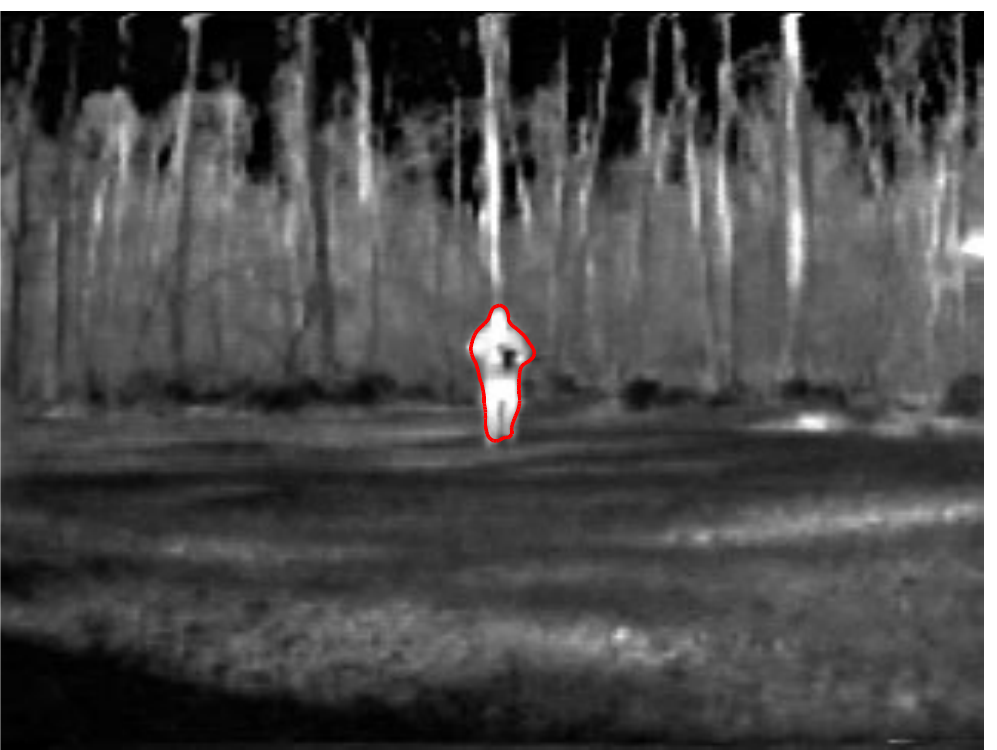,width=0.175\textwidth}
  \epsfig{figure=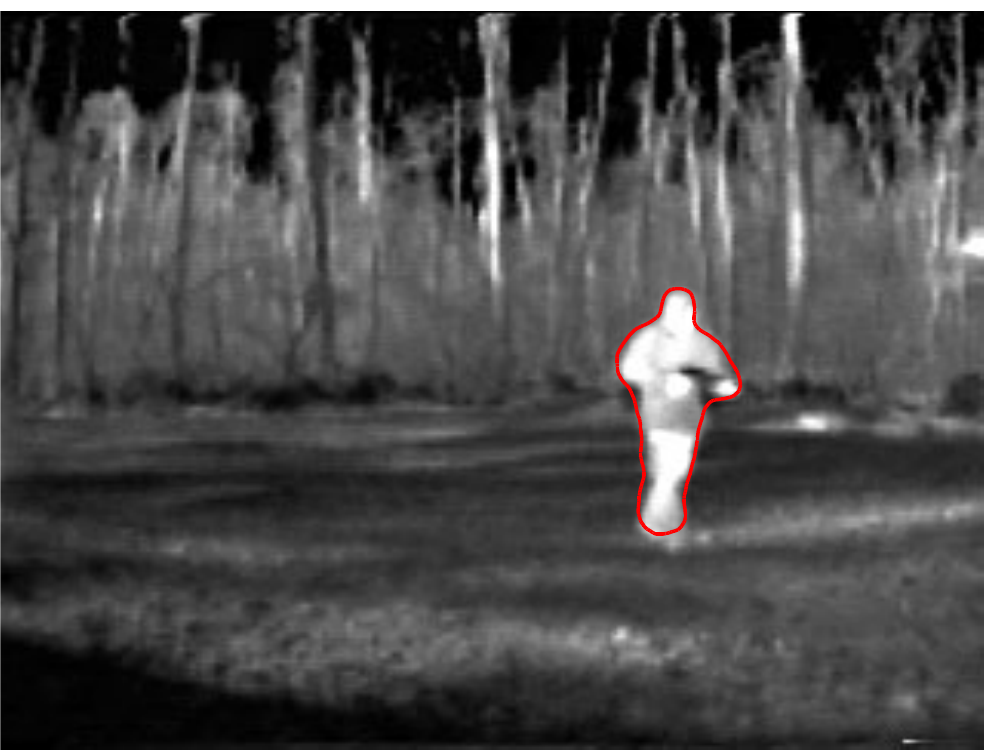,width=0.175\textwidth}
  \label{fig:exp3_irw07CVS}
}
\subfigure[E-SAc for Terravic video iruw02]{
  \epsfig{figure=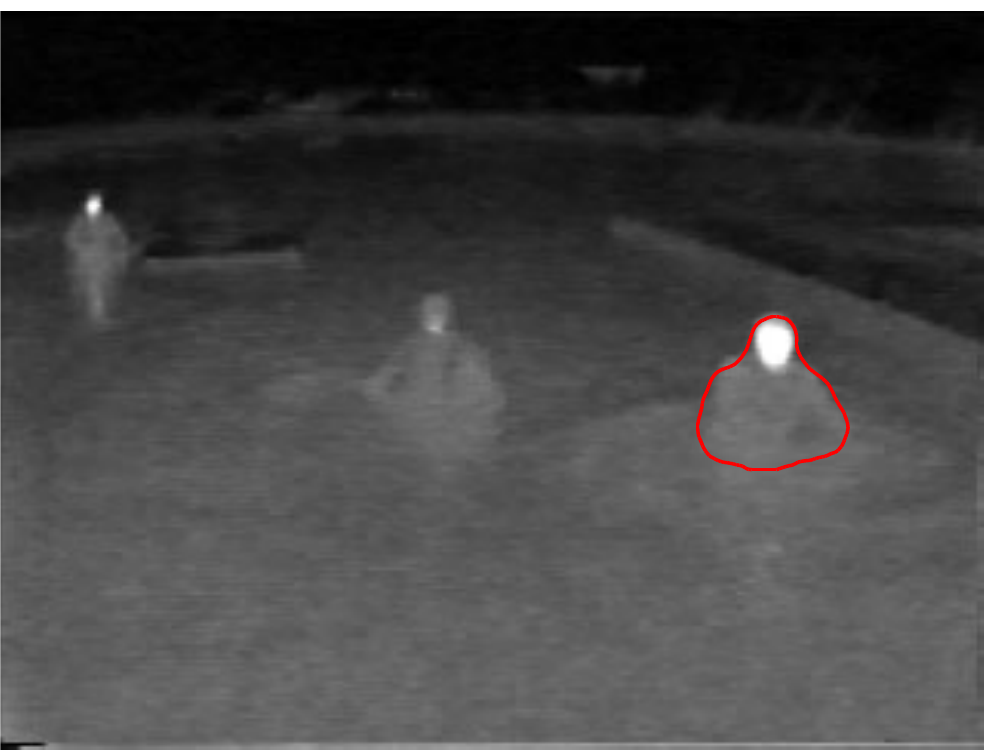,width=0.175\textwidth}
  \epsfig{figure=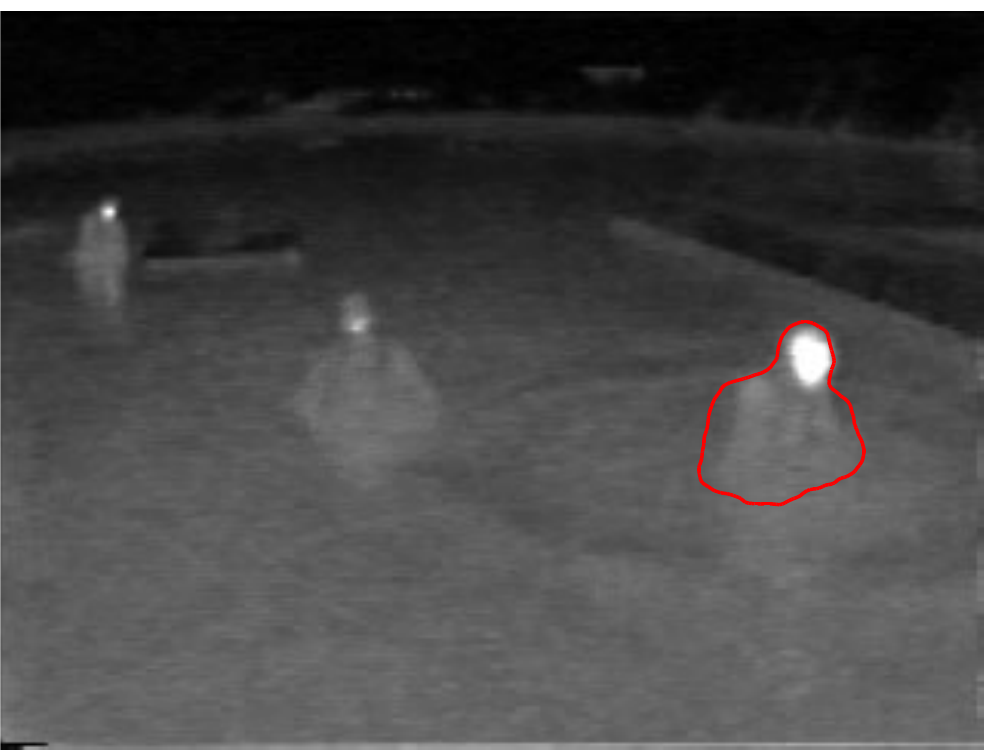,width=0.175\textwidth}
  \epsfig{figure=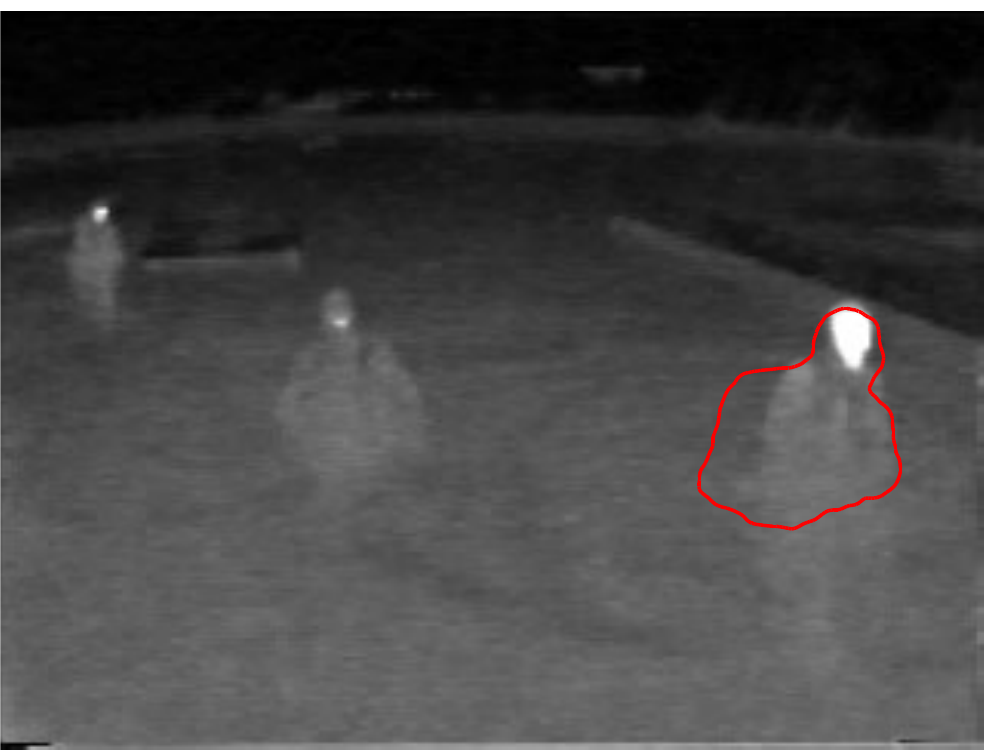,width=0.175\textwidth}
  \epsfig{figure=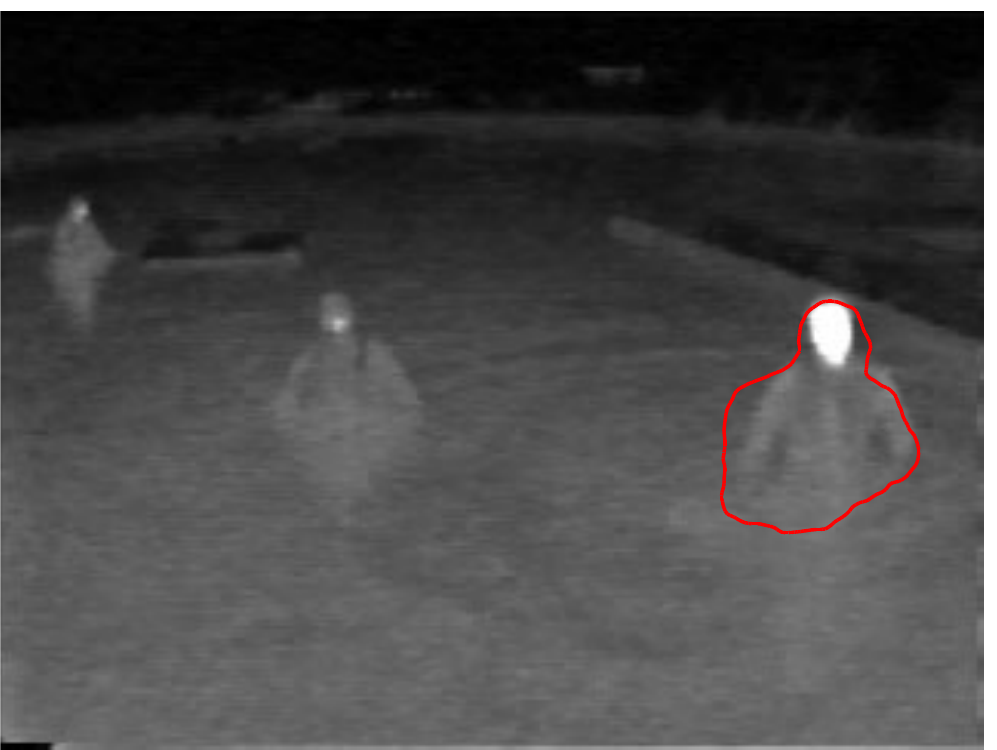,width=0.175\textwidth}
  \epsfig{figure=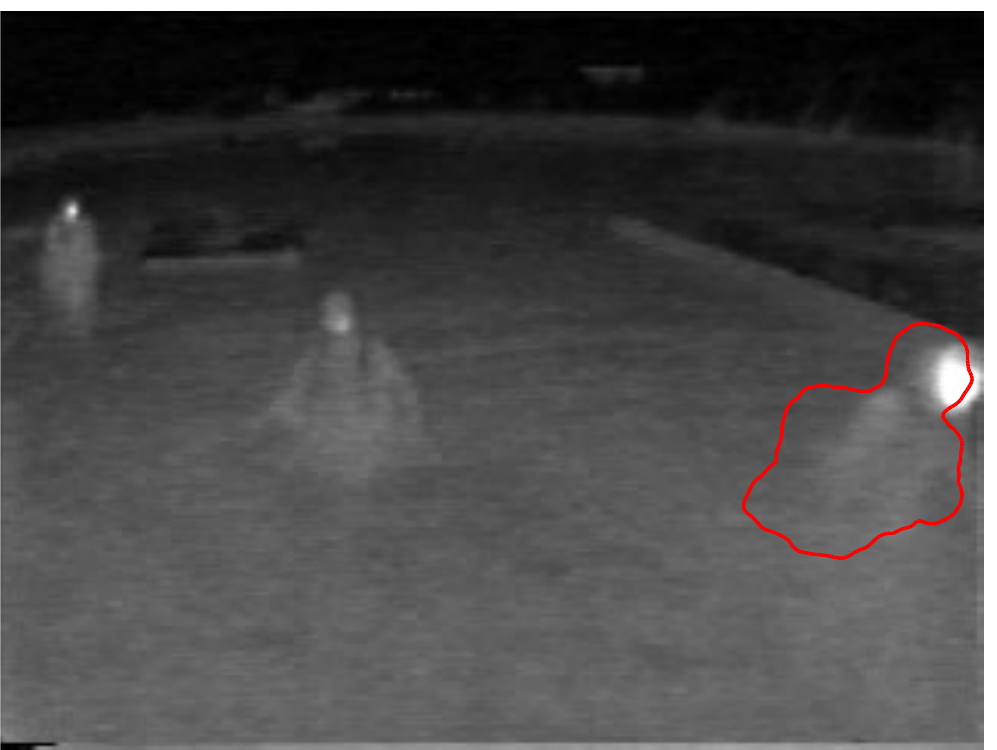,width=0.175\textwidth}
  \label{fig:exp3_iruw02ESA}
}

\subfigure[CV-S for Terravic video iruw02]{
  \epsfig{figure=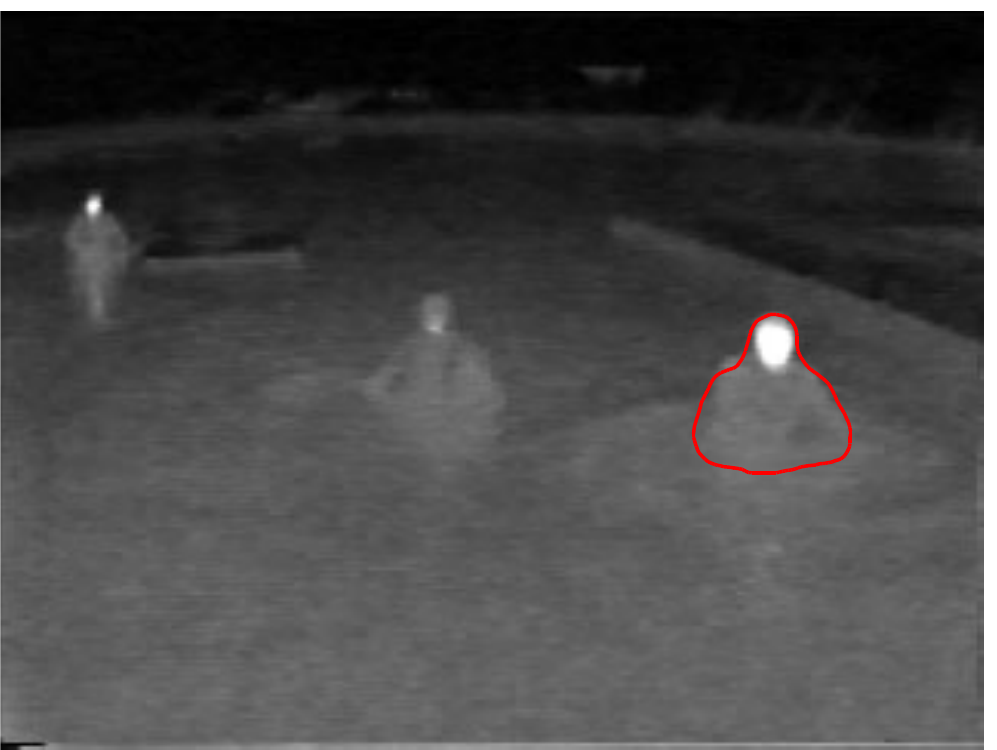,width=0.175\textwidth}
  \epsfig{figure=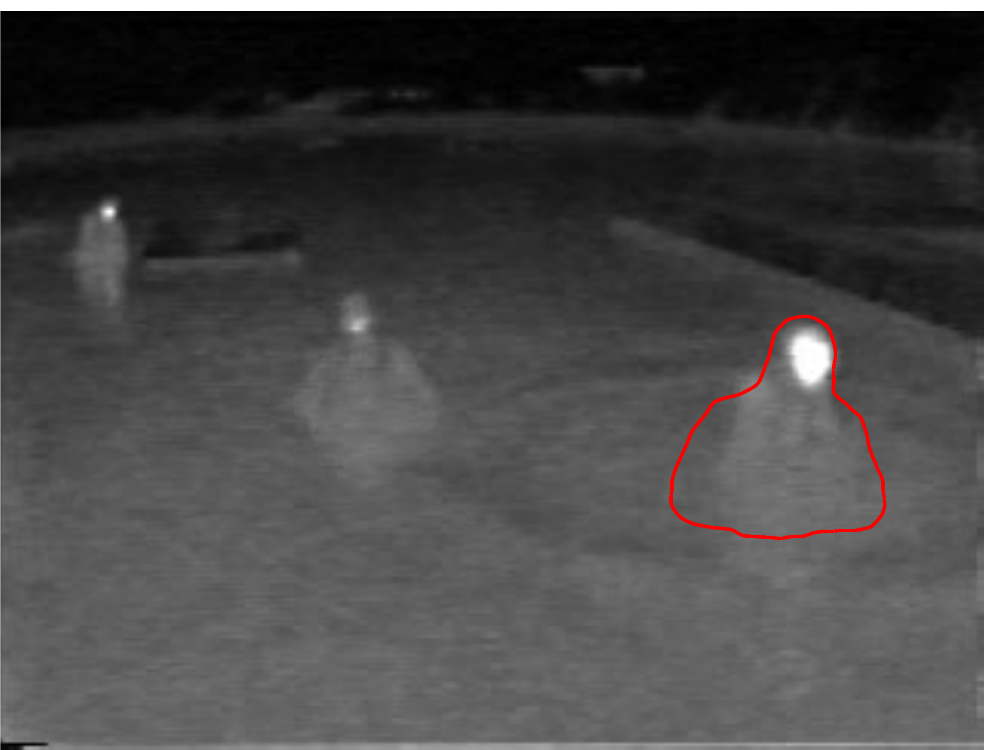,width=0.175\textwidth}
  \epsfig{figure=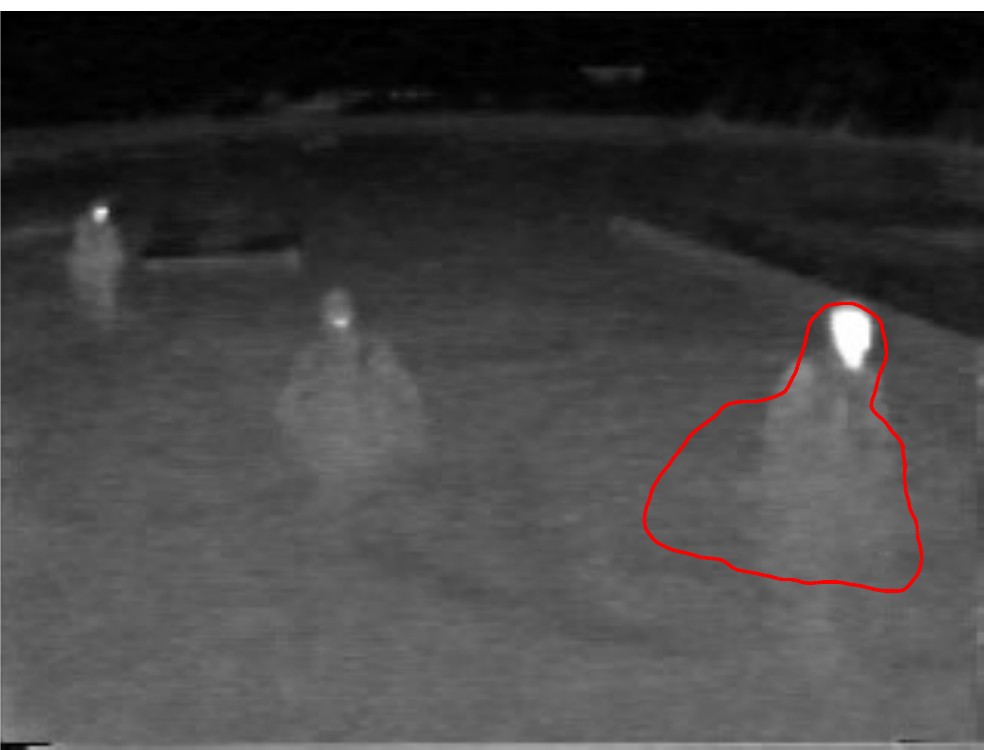,width=0.175\textwidth}
  \epsfig{figure=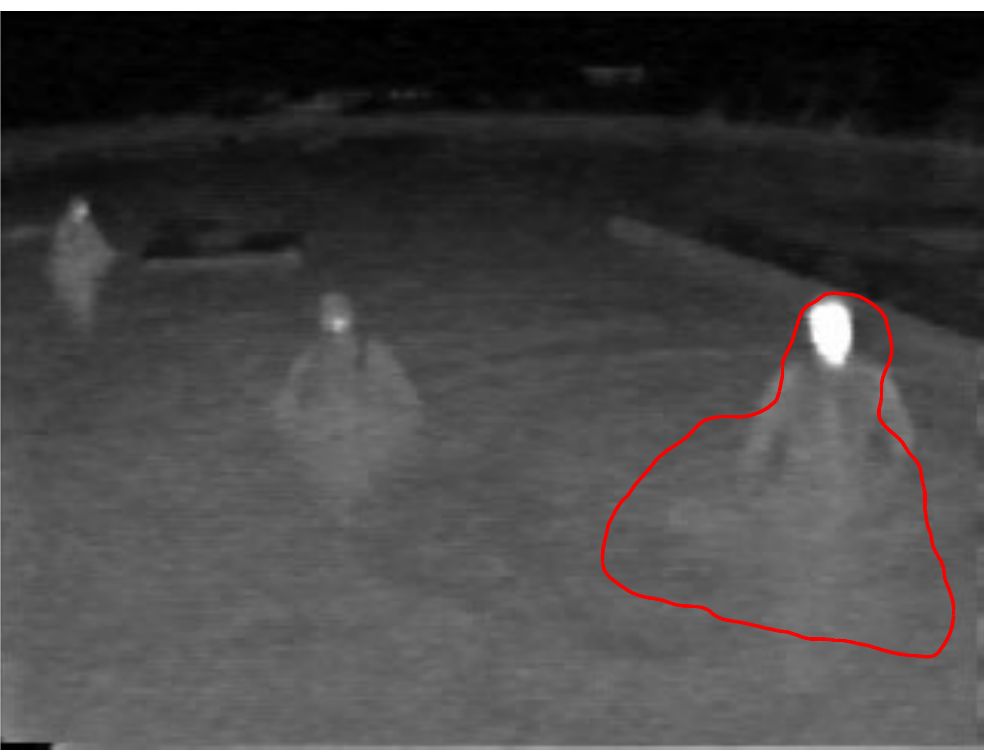,width=0.175\textwidth}
  \epsfig{figure=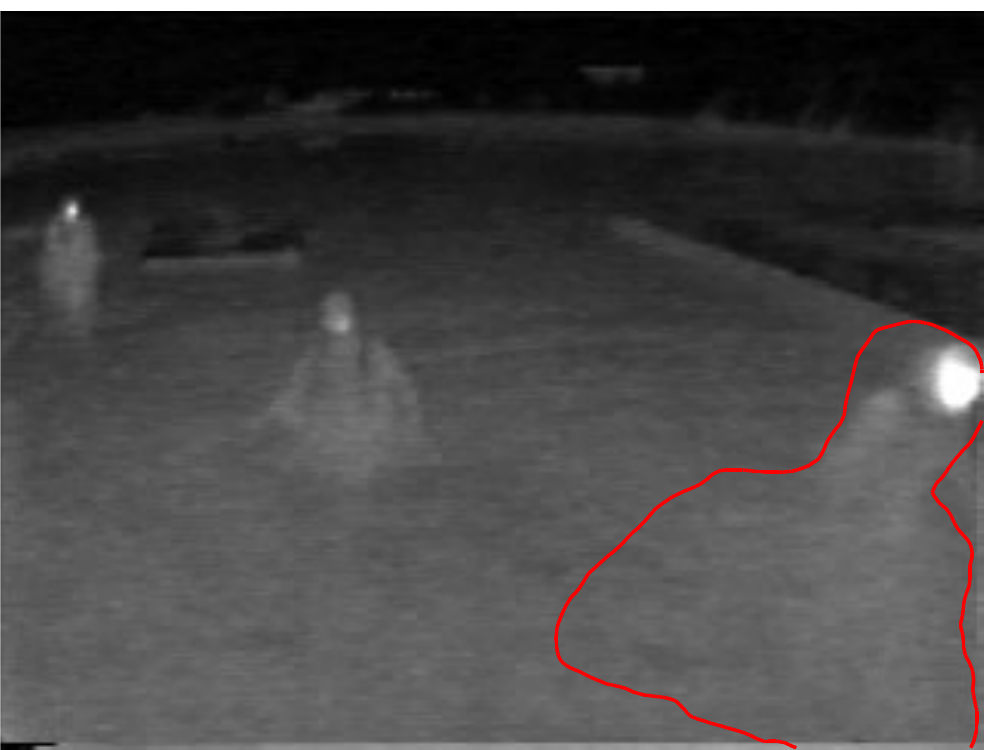,width=0.175\textwidth}
  \label{fig:exp3_iruw02CVS}
}
\subfigure[E-SAc for truck video]{
  \epsfig{figure=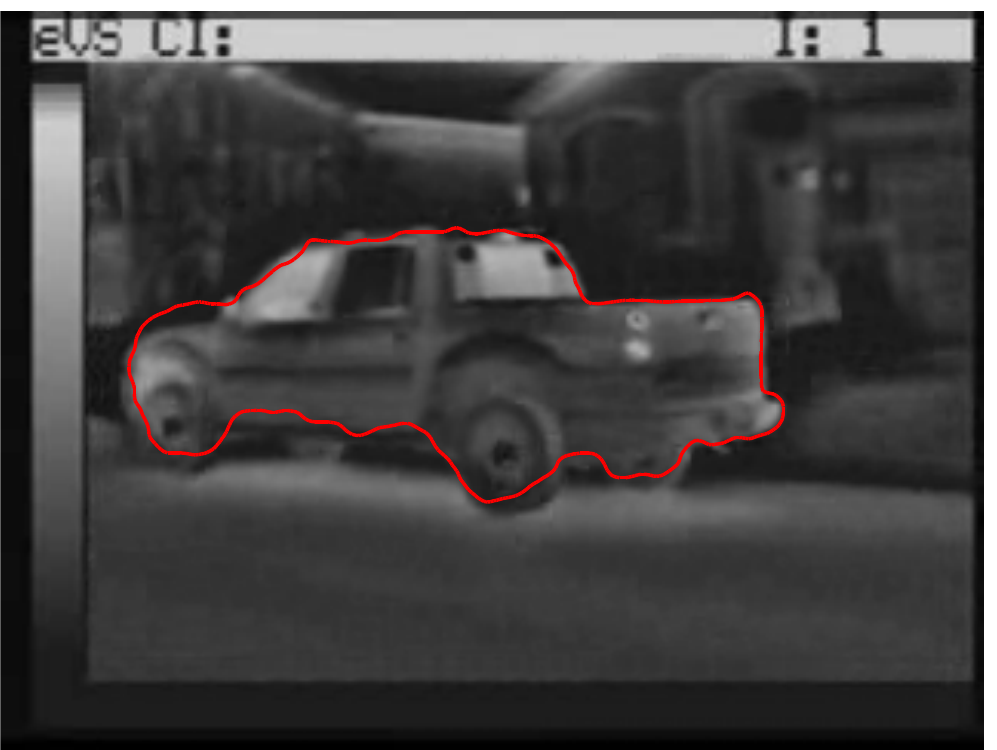,width=0.175\textwidth}
  \epsfig{figure=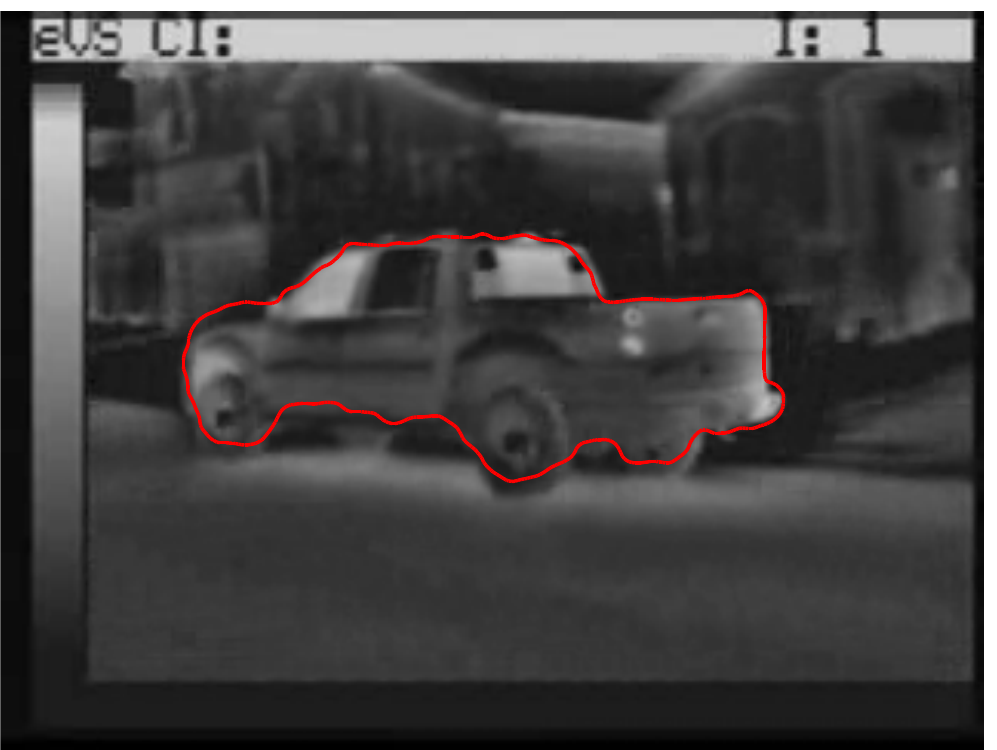,width=0.175\textwidth}
  \epsfig{figure=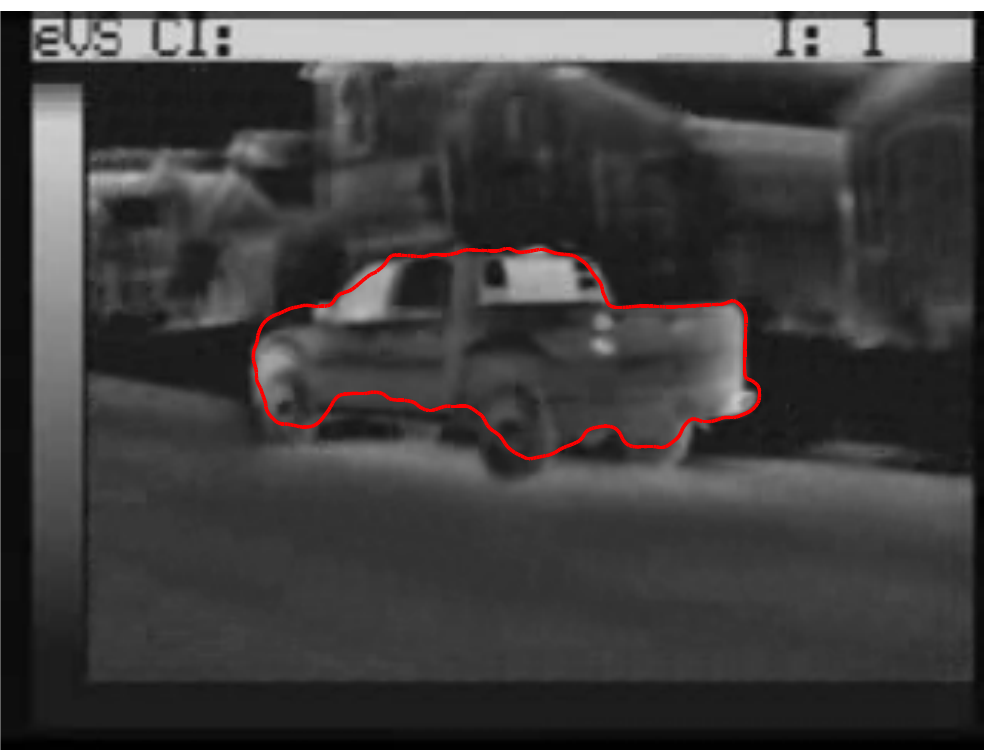,width=0.175\textwidth}
  \epsfig{figure=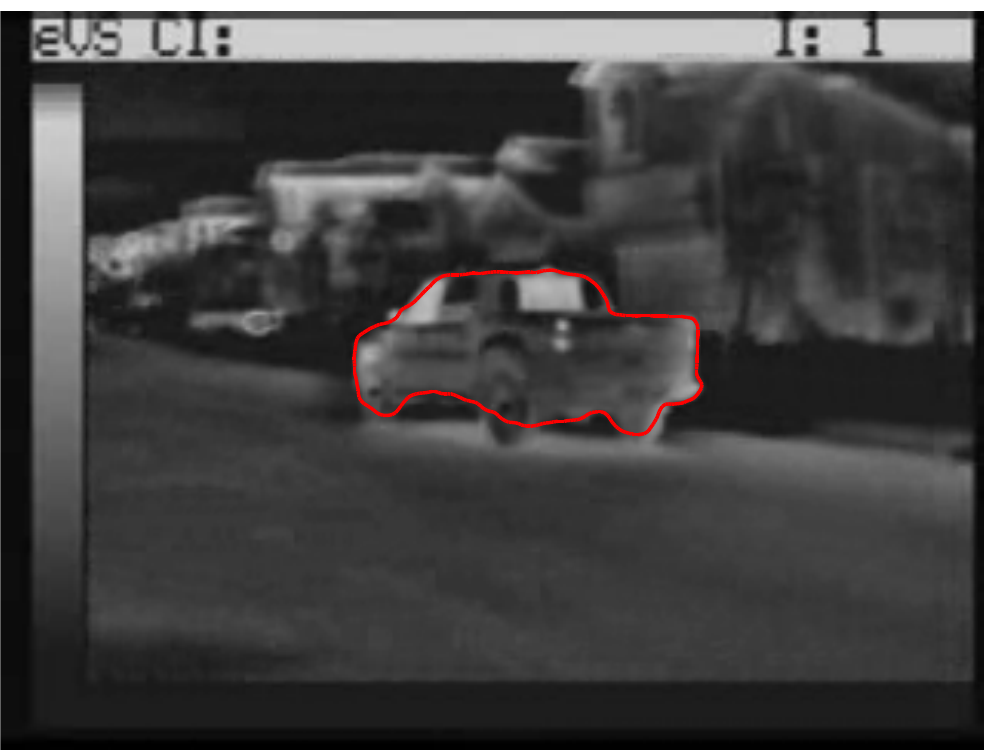,width=0.175\textwidth}
  \epsfig{figure=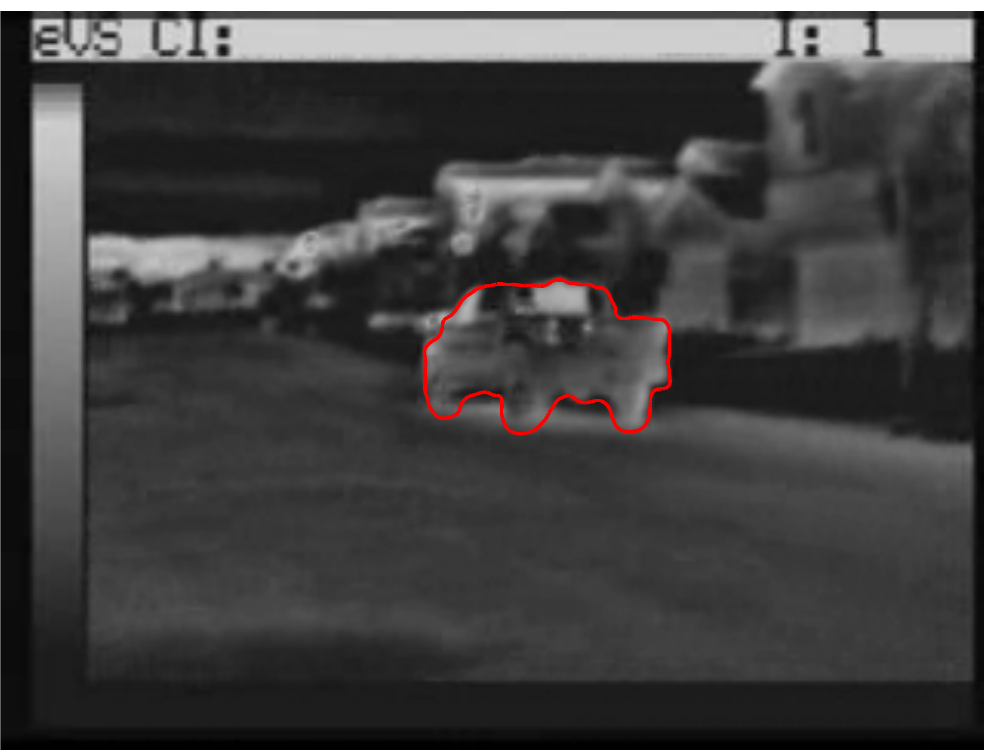,width=0.175\textwidth}
  \label{fig:exp3_itruckESA}
}
\subfigure[CV-S for truck video]{
  \epsfig{figure=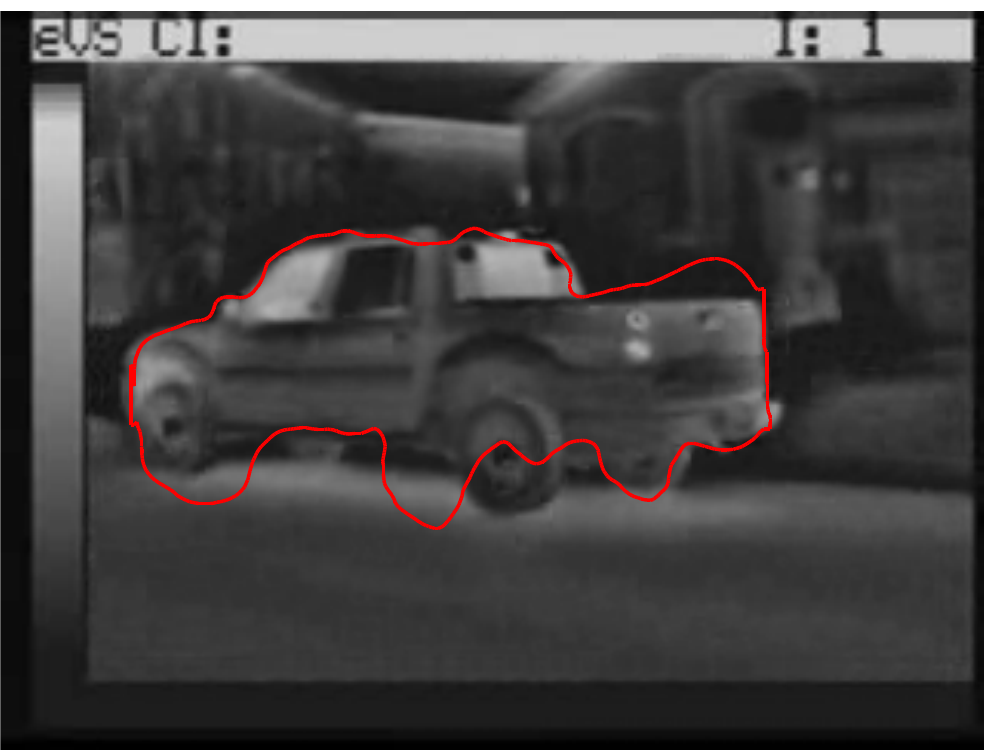,width=0.175\textwidth}
  \epsfig{figure=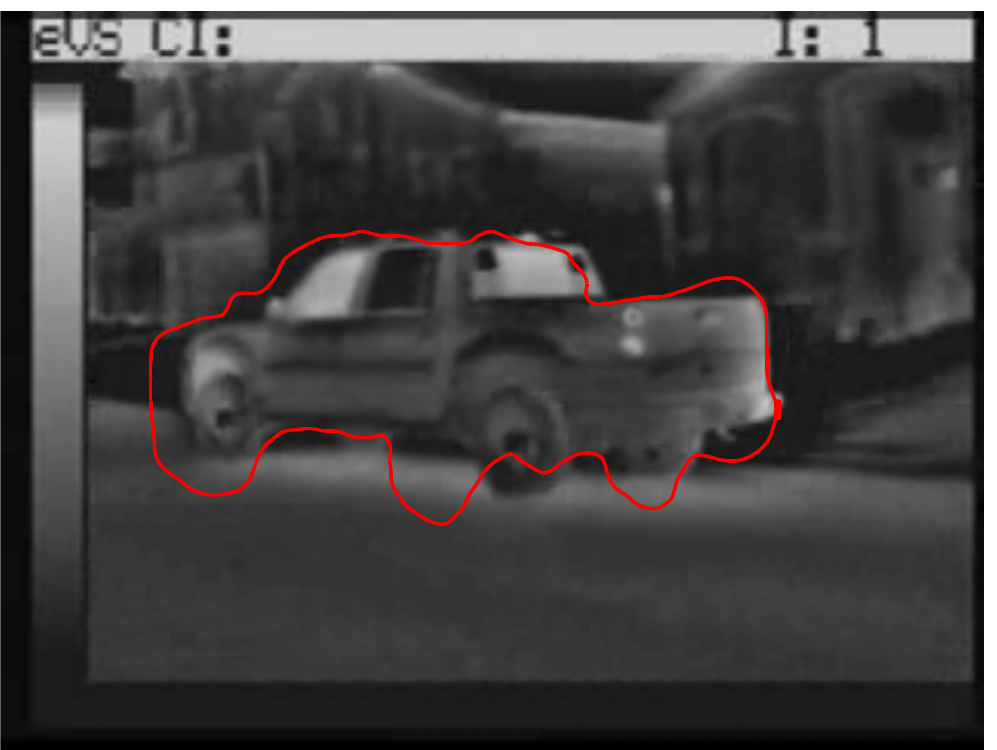,width=0.175\textwidth}
  \epsfig{figure=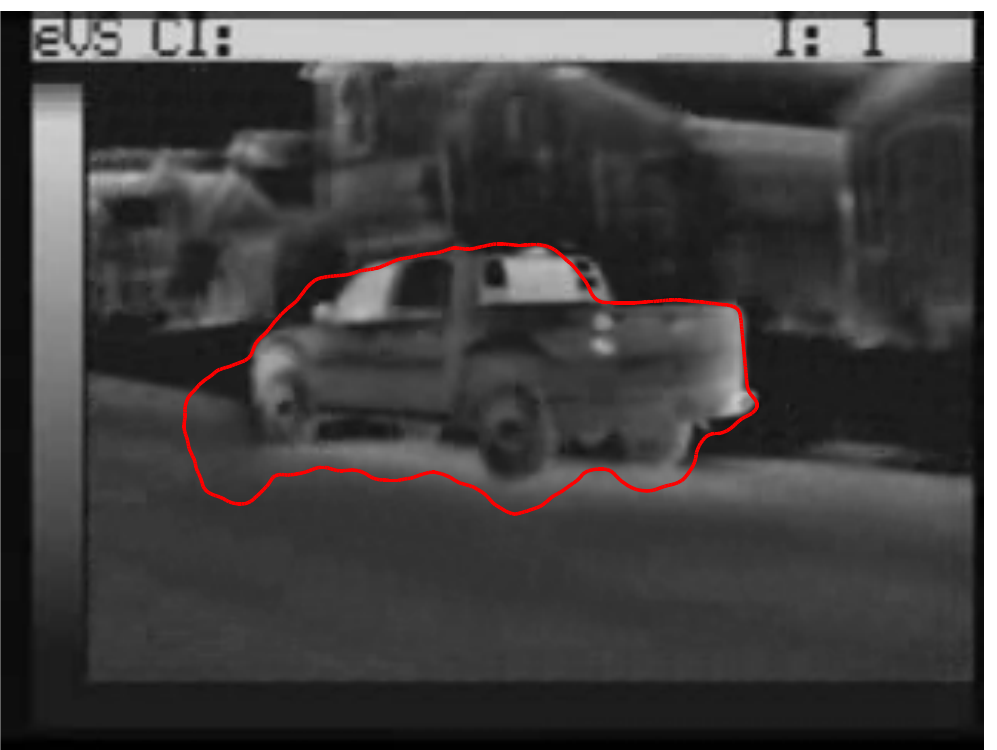,width=0.175\textwidth}
  \epsfig{figure=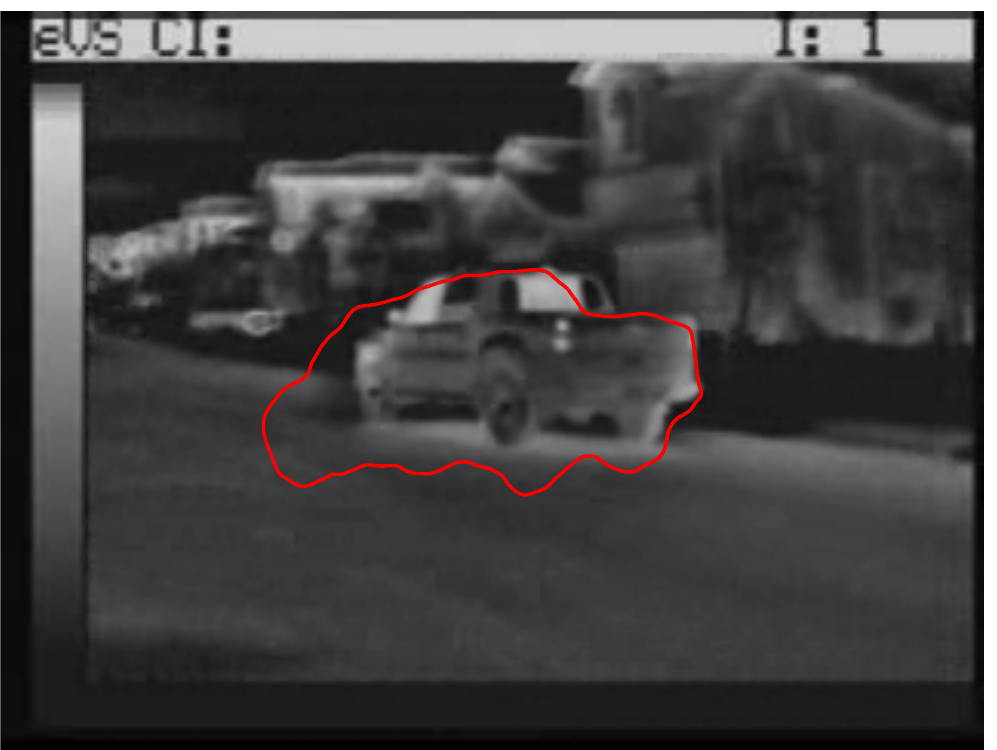,width=0.175\textwidth}
  \epsfig{figure=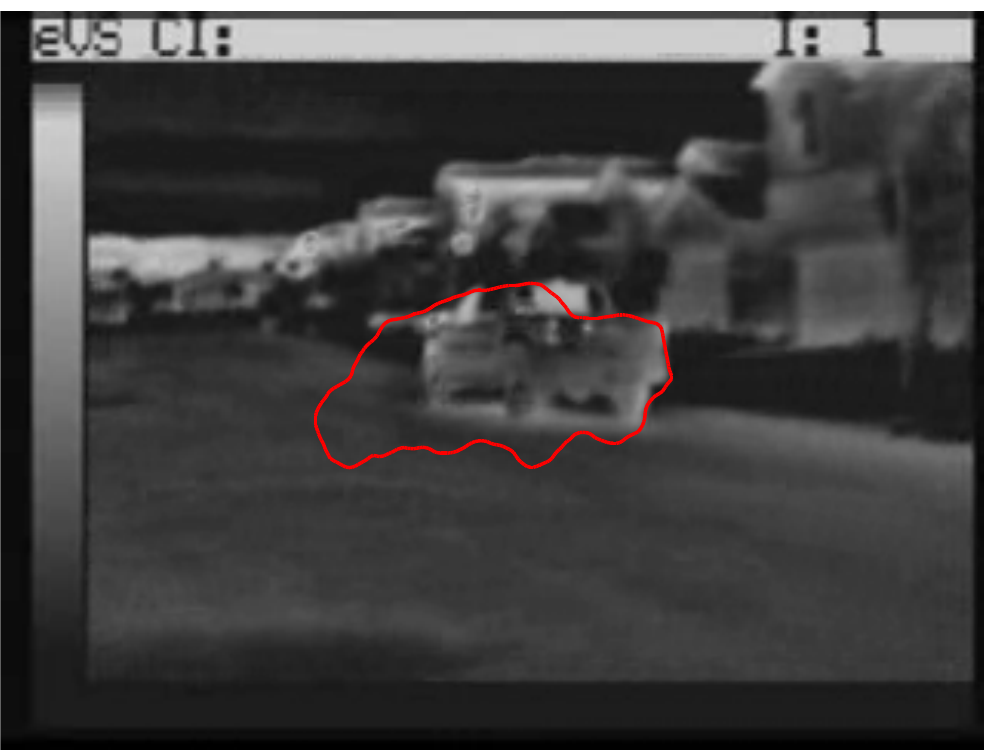,width=0.175\textwidth}
  \label{fig:exp3_itruckCVS}
}
\caption{Experiment 3 (Tracking): Tracking results on Terravic videos irw07, iruw02, and a truck video (Images by courtesy of \url{http://www.federalinfrared.com/}) showing several frames in each row. Incorporating object appearance in E-SA results in more accurate tracking in homogeneous (iruw02) or cluttered (truck) environments. }
\label{fig:exp3}
\end{figure*}

\medskip
\noindent \textbf{Experiment 3 (Tracking).} Next, our method is applied to tracking on infrared videos. Due to the smoothness of infrared images, vision tasks in this modality are particularly challenging for methods based on local features. Region-based active contour techniques---as the ones presented here---are, therefore, a viable alternative. In the following experiments, we demonstrate how E-SA (shape and appearance prior) improves tracking results compared to CV-S (shape priors only). Shape and appearance training models are first learned based on a few images from the sequence. In the tracking phase, the algorithms are performed with the initial guess in each frame being the final estimate of the previous frame.

The results on two videos from the Terravic data\-base~\cite{terravic} and one online video are shown in \autoref{fig:exp3}. For video irw07 in \autoref{fig:exp3_irw07ESA} and \autoref{fig:exp3_irw07CVS}, the training models are obtained from 12 frames of various videos from the Terravic database also including three frames from irw07. Since the person to be tracked is a very bright object with high contrast to the background, it is not surprising that both CV-S and E-SAc perform well. A more challenging scenario is shown for video iruw02 in \autoref{fig:exp3_iruw02ESA} and \autoref{fig:exp3_iruw02CVS}, since the contrast of the diver's body to the background is quite low. The CV-S and E-SAc training models are obtained from four frames of this video. E-SAc performs well in tracking the essentials of the shape. CV-S, on the other hand, diverges into the background. Finally, a video of a truck (by courtesy of \url{http://www.federalinfrared.com/}) with more appearance structure is shown in \autoref{fig:exp3_itruckESA} and \autoref{fig:exp3_itruckCVS}. Again, CV-S and E-SAc are trained on a few images of the scene. E-SAc provides very accurate tracking, whereas CV-S keeps track but not very accurately so. These experiments demonstrate how adding our one-dimensional appearance-based model may considerably increase robustness while being less complex than two-dimensional templates.

\section{Conclusion}
The concept of the photo-geometric representation of an object and its incorporation in a shape and appearance prior for active contour segmentation presents a compromise between complex appearance-based models involving two-dimensional templates and simple appearance-based models involving a finite number of statistics. Our experiments de\-mon\-strate how this coupling of shape and appearance leads to better accuracy compared to the latter methods for various applications. Whereas better accuracy compared to two-dimensional templates cannot be expected in general, our method has proved to be useful in the shown experiments and may thus be preferred due to its simpler representation of appearance. Our future research will focus on medical imaging applications of the photo-geometric representation for segmentation.

{\small
\bibliographystyle{./hieeetr.bst}
\bibliography{deformotionBib}
}

\end{document}